\documentclass{article}

\usepackage{arxiv}

\usepackage[utf8]{inputenc} 
\usepackage[T1]{fontenc}    
\usepackage{hyperref}       
\usepackage{url}            
\usepackage{booktabs}       
\usepackage{amsfonts}       
\usepackage{nicefrac}       
\usepackage{microtype}      
\usepackage{lipsum}		
\usepackage{graphicx}
\usepackage[numbers]{natbib}
\usepackage{doi}
\usepackage{subcaption}
\usepackage{xcolor}
\usepackage{framed,multirow}

\title{HDGL: A hierarchical dynamic graph representation learning model for brain disorder classification}

\author{Parniyan~Jalali\\
	Department of Electrical and Computer Engineering\\
	Isfahan University of Technology\\
	Isfahan, Iran \\
	\texttt{p.jalali@ec.iut.ac.ir} \\
	\And
	 Mehran~Safayani\\
	Department of Electrical and Computer Engineering\\
	Isfahan University of Technology\\
	Isfahan, Iran \\
	\texttt{safayani@iut.ac.ir} \\
}

\date{}

\renewcommand{\headeright}{}
\renewcommand{\undertitle}{}
\renewcommand{\shorttitle}{}

\hypersetup{
pdftitle={HDGL},
pdfsubject={q-bio.NC, q-bio.QM},
pdfauthor={Parniyan~Jalali, Mehran~Safayani},
pdfkeywords={Graph classification, Dynamic functional connectivity, Hierarchical model, Non-imaging information, Graph representation learning, Spatial-temporal modeling}
}

\begin{document}
\maketitle

\begin{abstract}
The human brain can be considered as complex networks, composed of various regions that continuously exchange their information with each other, forming the brain network graph, from which nodes and edges are extracted using resting-state functional magnetic resonance imaging (rs-fMRI).
Therefore, this graph can potentially depict abnormal patterns that have emerged under the influence of brain disorders. So far, numerous studies have attempted to find embeddings for brain network graphs and subsequently classify samples with brain disorders from healthy ones, which include limitations such as: not
considering the relationship between samples, not
utilizing phenotype information, lack
of  temporal analysis,
using
static functional connectivity (FC)
instead of dynamic ones and using a fixed graph structure.
We propose a hierarchical dynamic graph representation learning (HDGL) model, which is the first model designed to address all the aforementioned challenges. HDGL consists of two levels, where at the first level, it constructs brain network graphs and learns their spatial and temporal embeddings, and at the second level, it forms population graphs and performs classification after embedding learning.
Furthermore, based on how these two levels are trained, four methods have been introduced, some of which are suggested for reducing memory complexity.
We evaluated the performance of the proposed model on the ABIDE and ADHD-200 datasets, and the results indicate the improvement of this model compared to several state-of-the-art models in terms of various evaluation metrics.

\end{abstract}


\section{Introduction}
Machine learning techniques, applied extensively in various domains, particularly in classification task, aim to train models for accurate data point (e.g., images, text and graphs) labeling. Enhancing the classifier's performance is often achieved through the acquisition of more informative data representations \citep{alanazi2017critical,JIANG2020104096}.
Graphs are useful tools for modeling complex relationships between different entities. In this way, each node represents an entity, and the edges represent the relationships between them. Many real-world phenomena, such as citation networks \citep{pan2019learning}, transportation networks \citep{zhang2020spatio,tang2022spatio}, molecular structures \citep{hy2023multiresolution}, and also brain networks \citep{zhang2022diffusion}, are modeled using graphs.
Therefore, extensive efforts have been made towards embedding graphs into meaningful representations and classifying them for various applications.

The human brain can be regarded as a complex network composed of various regions. Each region, while performing its specific function, regularly exchanges information with other regions, forming a complex network known as the brain network. Functional Magnetic Resonance Imaging (fMRI) is a non-invasive neuroimaging technique that captures 4D fMRI data as a series of three-dimensional volumes over time. This imaging method allows us to assess brain function by detecting changes in blood flow and oxygenation levels. Each volume consists of voxels, which are the smallest measurable elements in fMRI images, and each voxel contains the Blood Oxygen Level Dependent (BOLD) signal \citep{smith2004overview}. fMRI provides both spatial information about brain structure, connectivity between its regions, and temporal information about regional activity.
The brain network can be represented as a graph, where nodes represent brain region of interests (ROIs), and edges depict the inter-regional functional associations extracted from fMRI data \citep{liu2023braintgl}.

The brain network illustrates inter-regional interactions and reveals specific abnormal patterns in the presence of brain disorders \citep{saeidi2022decoding}. The aim of this research is to find an appropriate representation for brain network graphs and subsequently classify healthy samples from those with brain disorders such as autism spectrum disorder (ASD) and attention deficit hyperactivity disorder (ADHD).

Various studies have been conducted on graph embedding of brain network graphs and their classification.  However, some of these studies exhibit certain limitations in one or more aspects, as outlined below: 1) Considering each sample individually: this is while recent research in the field of disease diagnosis \citep{parisot2017spectral,JIANG2020104096,kazi2019inceptiongcn} shows that considering the relationships between similar samples and utilizing their information can improve the model's performance; 2) Not utilizing phenotype information: typically, the use of phenotype information such as age, gender, site, etc., along with imaging data can lead to improved model performance\citep{zhao2022dynamic}; 3) Lack of  temporal analysis: most studies \citep{JIANG2020104096,zaripova2023graph,eslami2019asd} have only focused on spatial analysis and have overlooked the temporal analysis of ROI-time series; 4) Utilizing static functional connectivity (FC): this is while a recent research \citep{lurie2020questions} has shown that the statistical dependency between ROI-time series changes over time; 5) Using a fixed graph structure: the initial graph structure may contain noise and incorrect connections, which can disrupt the message-passing process and significantly impact the performance of Graph Neural Networks (GNNs), resulting in inaccurate node embeddings \citep{kazi2022differentiable}.
 
The above concerns motivate us to propose a hierarchical dynamic graph representation learning (HDGL) model, which is the first model designed to address all the aforementioned challenges. In order to address the first and second challenges, a hierarchical framework consisting of two levels has been proposed, which utilizes brain network graphs at the first level and population graphs at the second level. The nodes of the population graph represent the subjects, and the edge weights are assigned based on measuring the similarity between the phenotypic information of the subjects. Therefore, by constructing a population graph, the dependencies between the subjects are modeled, and in addition to using imaging data, the phenotypic information is used for the classification. To address the third issue, temporal analysis has been performed on the ROI-time series using Gated Recurrent Unit (GRU), and the resulting representations are used as node features. To address the fourth issue, the proposed model employs the sliding window technique to extract dynamic FC and subsequently construct graphs over time. It then utilizes a Transformer to learn temporal embeddings of these graphs. In response to the fifth problem, the Self Attention Graph Pooling (SAGPool) method has been employed. In addition to reducing the input dimension and preventing overfitting, this method is capable of preserving important nodes and connections during training while discarding the rest. Therefore, ultimately, a more appropriate structure is obtained for the initial graphs.

Additionally, the proposed model is trained using both transductive and inductive approaches, and the results are examined. In the transductive setting, the first and second levels of the model are trained either separately or end-to-end. In the end-to-end training mode, the model requires a significant amount of memory during execution. To address this issue, a scalable algorithm has been introduced, which notably reduces the memory requirement. 
Additionally, in the inductive setting, both levels are trained in an end-to-end and supervised manner.
Overall, the contributions of our work are summarized as
below:
\begin{itemize}
	\item
	A novel hierarchical framework named HDGL is proposed for classifying brain disorders with resting state functional MRI (rs-fMRI). It employs a sliding window technique to extract dynamic FC and uses the GRU to perform temporal analysis on ROI-time series. After learning brain graph embeddings and identifying biomarkers, it then uses a population graph to incorporate non-imaging information and the relationships between subjects into the framework.
	\item
	We examined different approaches for training the model, and in this regard, we proposed a scalable algorithm that trains in semi-supervised manner and reduces memory requirements.
	\item
	The experimental results on two public medical datasets
	for ASD and ADHD classification demonstrate that the proposed
	HDGL achieves state-of-the-art performance.
\end{itemize}
\begin{figure*}[!h]
	\begin{center}
		\includegraphics[scale=.5]{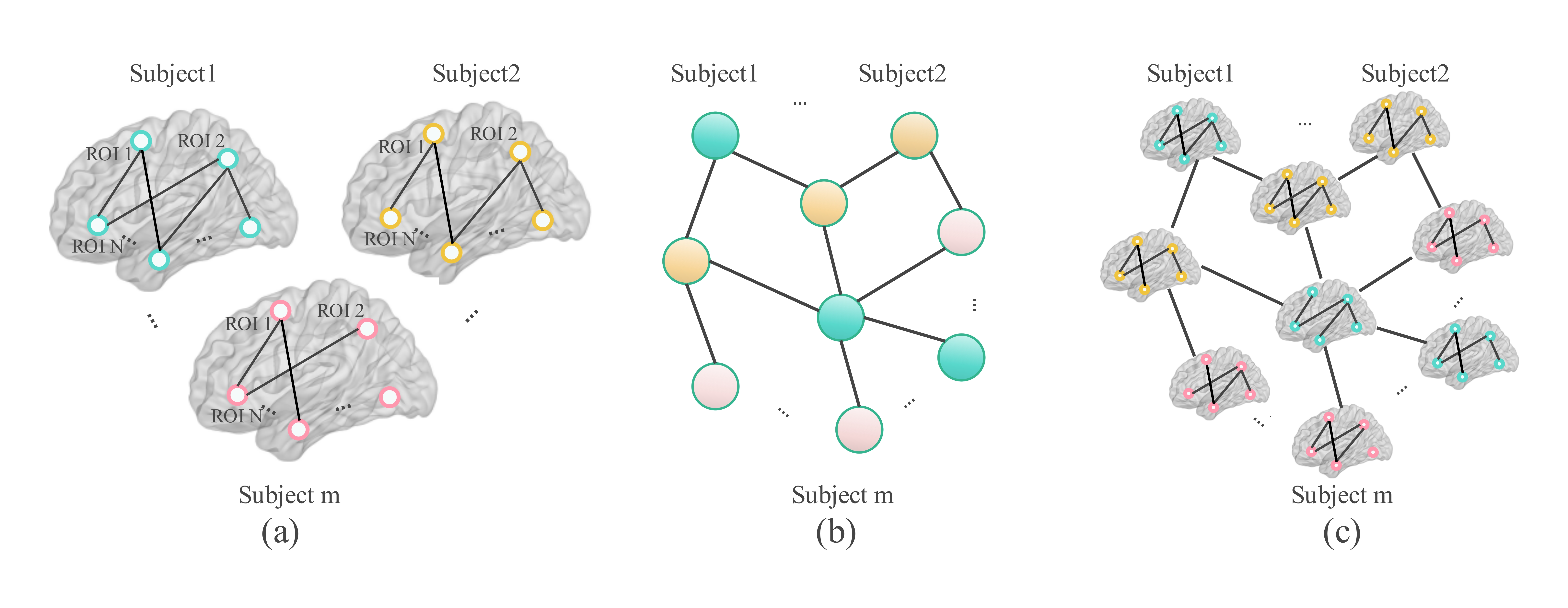}
		\caption{\label{fig0}Illustration of various types of graphs defined in brain disorder classification methods, assuming we have $ m $ subjects and $ N $ ROIs: (a) Brain-network based approaches (b) population-graph based approaches (c) hierarchical-graph based approaches  }	
	\end{center}	
\end{figure*}

The remainder of this paper is organized as follows: In Section \ref{section2}, we discuss related works on brain disorder classification. Section \ref{section3} introduces the HDGL model and its various components, along with an explanation of four different training methods. The performance of various types of HDGL models has been evaluated on the ABIDE and ADHD-200 datasets in Section \ref{section4}. Section \ref{section5} presents various ablation studies along with the analysis of their results. Finally, we discuss the limitations and future work in Section \ref{section6}, and conclude this work in Section \ref{section7}.

\section{Related Works}
\label{sec:headings}
\label{section2}
Recent research studies have made significant efforts towards modeling the rs-fMRI data in the form of graphs. These studies are categorized into three groups based on the type of defined graphs: Brain-network based approaches, population-graph based approaches, and hierarchical-graph based approaches.

\subsection{Brain-network based}
In these methods, as shown in Fig \ref{fig0}(a), each brain region is considered as a node, and the edges are determined through the analysis of time series data. Then, they attempt to extract informative representations and thus formulate the problem of classifying brain disorders as a graph classification task. To be more precise, these methods extract FC from the time series data and then consider the resulting matrix as the brain network's adjacency matrix after applying a threshold to its values. These methods can be classified into two categories based on the type of FC. Static FC and dynamic FC. 

Static FC methods extract the FC matrix from the ROI-time series across the entire imaging session. Thus, for each subject, we have a FC matrix.
Yan et al. \citep{yan2019groupinn} tackled the issues arising from noisy time series data obtained from fMRI, which might lead to inaccurate graphs. Additionally, they addressed the challenge of overfitting due to the high dimensionality of fMRI data with limited samples. To overcome these challenges, they introduced a node classification layer to reduce dimensionality and preserve important connections between super nodes while concealing noisy ones within each cluster. The proposed model then passed the graphs through multiple graph convolutional network (GCN) layers before feeding them into a fully connected layer for classification, resulting in a more accurate structure with reduced dimensionality for the initial brain graphs during training. Eslami et al. \citep{eslami2019asd} aim to reduce the input feature size by preserving only the correlations with the highest values and discarding the rest. Then, they use an autoencoder to extract a lower-dimensional representation for the features, followed by a single-layer perceptron for classification. Both introduced methods consider the brain graph as a spatial graph, meaning that no temporal analysis is performed on the ROI-time series. To address this issue, Azevedo et al. \citep{azevedo2022deep} attempted to propose a model that can perform both spatial and temporal analysis. The introduced model is composed of multiple dilated convolutional layers \citep{yu2015multi}, which are a specific type of one-dimensional convolution, for temporal analysis, and then multiple graph network blocks for spatial analysis of the brain graph. 
The methods introduced so far, by considering static FC, miss important information arising from the changes in brain activity states over time.

\begin{figure*}[!h]
	\begin{center}
		\includegraphics[scale=.5]{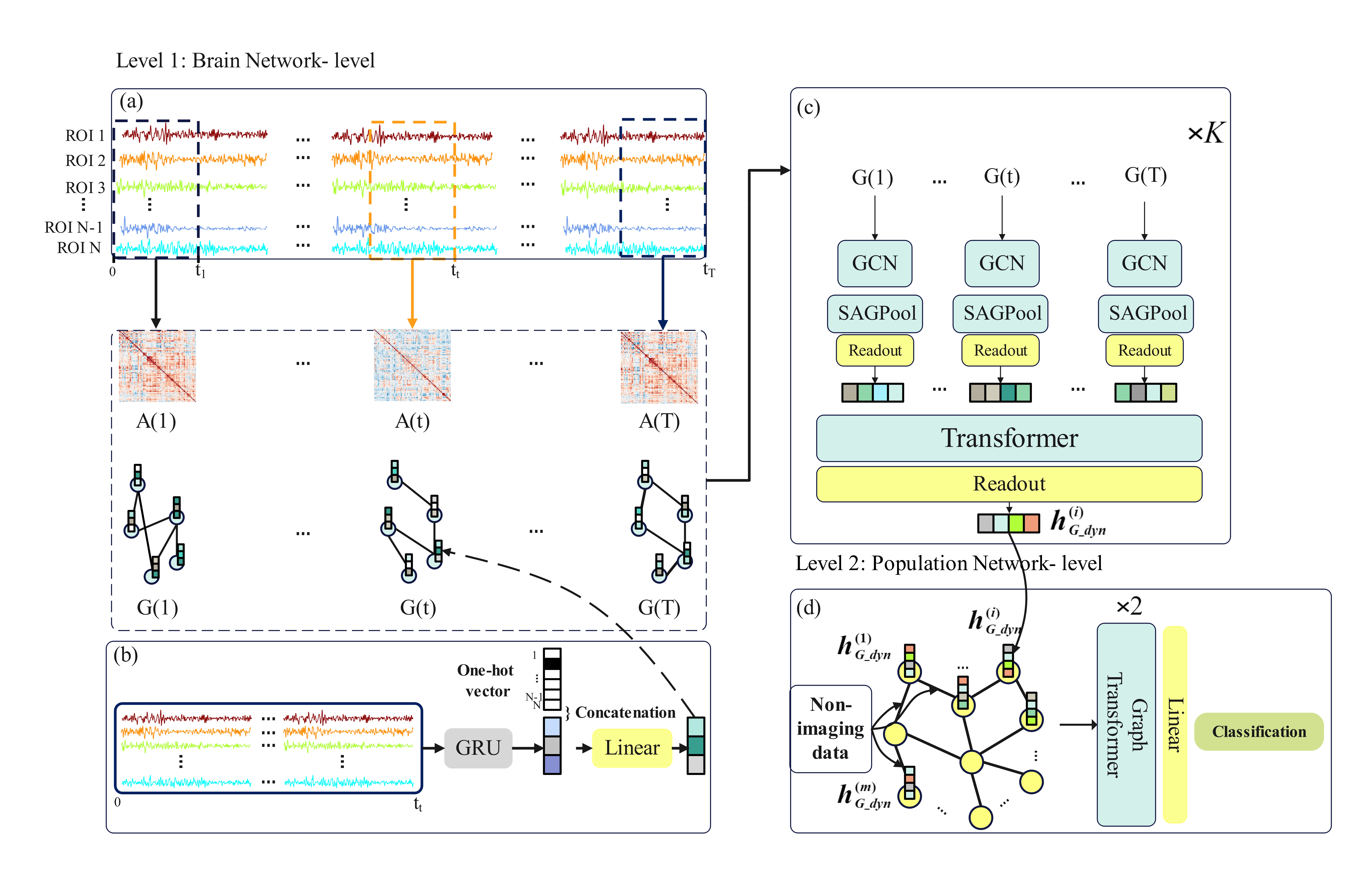}
		\caption{\label{fig1}HDGL architecture consists of four parts: (a) Adjacency matrix construction (b) Node feature extraction (c) Brain network analysis (d) Population network analysis. From another perspective, HDGL consists of two levels: the Brain network-level, which constructs the brain network graphs and learns embeddings for them, and the Population network-level, which constructs the population graph, learns embeddings for its nodes, and then classifies them.  }	
	\end{center}	
\end{figure*}

Dynamic FC methods, on the other hand, divide the ROI-time series into smaller time segments using sliding windows. Within each segment, the FC is extracted. As a result, for each subject, we have multiple FC matrices over time. Kim et al. \citep{kim2021learning} proposed a model capable of learning spatial and temporal embeddings of dynamic brain graphs. GNNs were employed for learning spatial embeddings, and subsequently, a readout model with an attention mechanism \citep{hu2018squeeze} was used to integrate representations of different graph nodes. Then, a Transformer was utilized for learning temporal embeddings. The obtained representation was fed into a single-layer feedforward network for classification.
Similar to the previous study, Liu et al. \citep{liu2023braintgl}also consider brain networks as dynamic graphs.  Then, GCNs along with an attention-based pooling method are utilized to correct the graph structures during the training process. Following that, a two-level temporal analysis is conducted on the graphs. At the first level, the ROI-time series  are analyzed using one-dimensional convolutions. Then, at the second level, sequences of brain graphs are fed into an Long Short-Term Memory (LSTM) model to perform temporal analysis at the graph level. The representation obtained from this stage is then fed into a fully connected layer for classification.
The results of these two studies indicate that modeling brain graphs as dynamic graphs, performing spatial and temporal analysis on these data, and correcting graph structures during training significantly improve the model's performance.

\subsection{Population-graph based}
According to Fig \ref{fig0}(b), in population graphs, each node represents a subject, and the edges are determined based on the similarity between the phenotypic and imaging data.
For instance, Parisot et al. \citep{parisot2017spectral}  utilize a static FC to define the features of population graph nodes and employ recursive feature elimination (RFE) as a dimensionality reduction technique. After that a multi-layer graph convolutional neural network based on ChebNet spectrum is employed, along with the Rectified Linear Unit (ReLU) activation function, to obtain node representations.
In this study, the edge weights of the population graph were obtained manually using similarity metrics. However, in a recent research \citep{huang2022disease}, a module called "pairwise association encoder (PAE)" was introduced, which takes non-image features of paired samples, passes them through two linear layers, and then calculates the similarity between two representations using cosine similarity.
One of the limitations of these methods is the separation of the feature extraction process from the training process. Additionally, considering a static FC and ignoring temporal variations are other challenges of these approaches.

\begin{figure}[!htb]
   \begin{minipage}{0.48\textwidth}
     \centering
     \includegraphics[width=.95\linewidth]{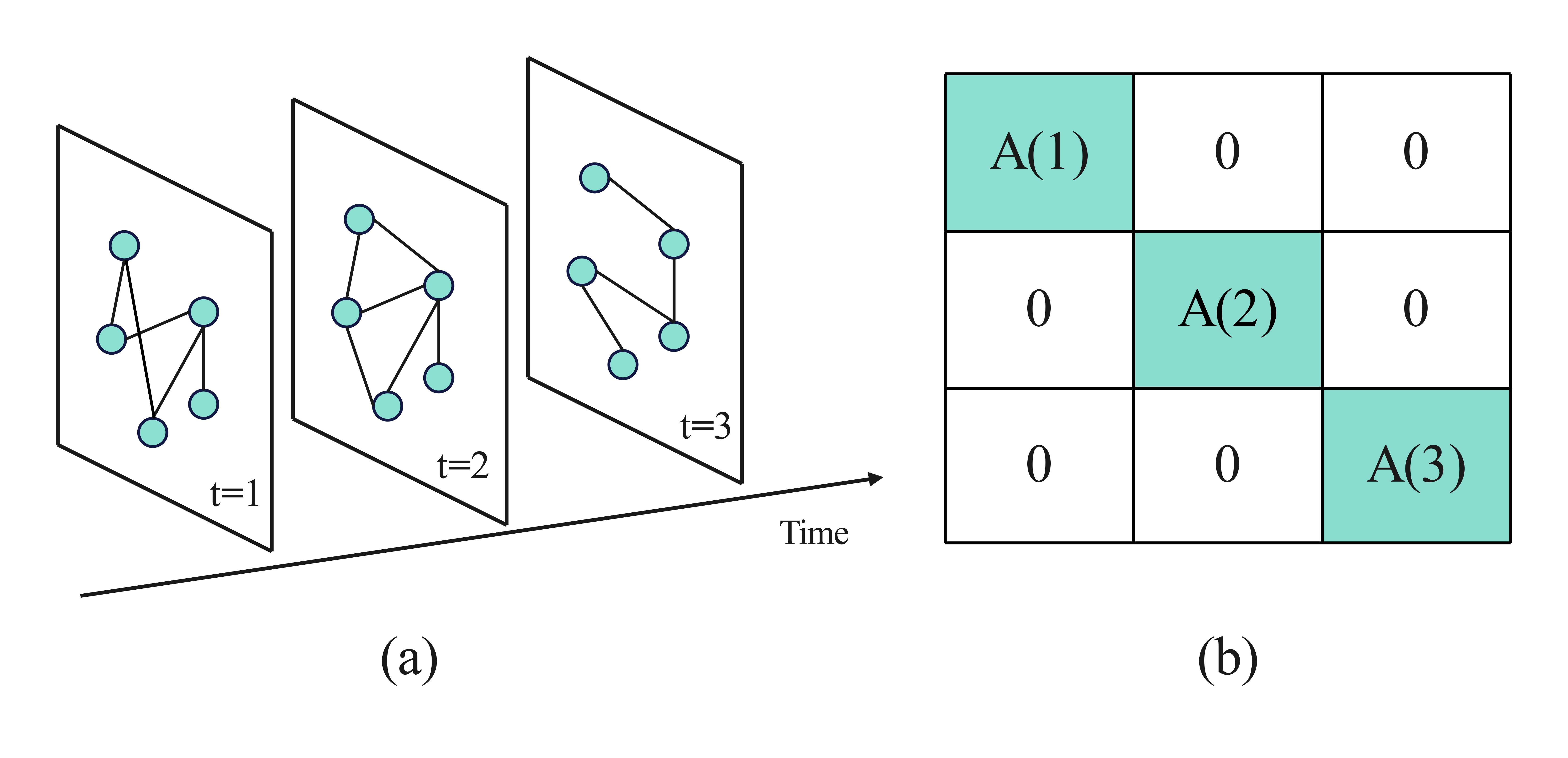}
     \caption{(a) The dynamic brain graph obtained for the  $ i $-th  subject under the assumption of $ T $=3. (b) Adjacency matrix of dynamic brain graph.}\label{fig2}
   \end{minipage}\hfill
   \begin{minipage}{0.48\textwidth}
     \centering
     \includegraphics[width=1\linewidth]{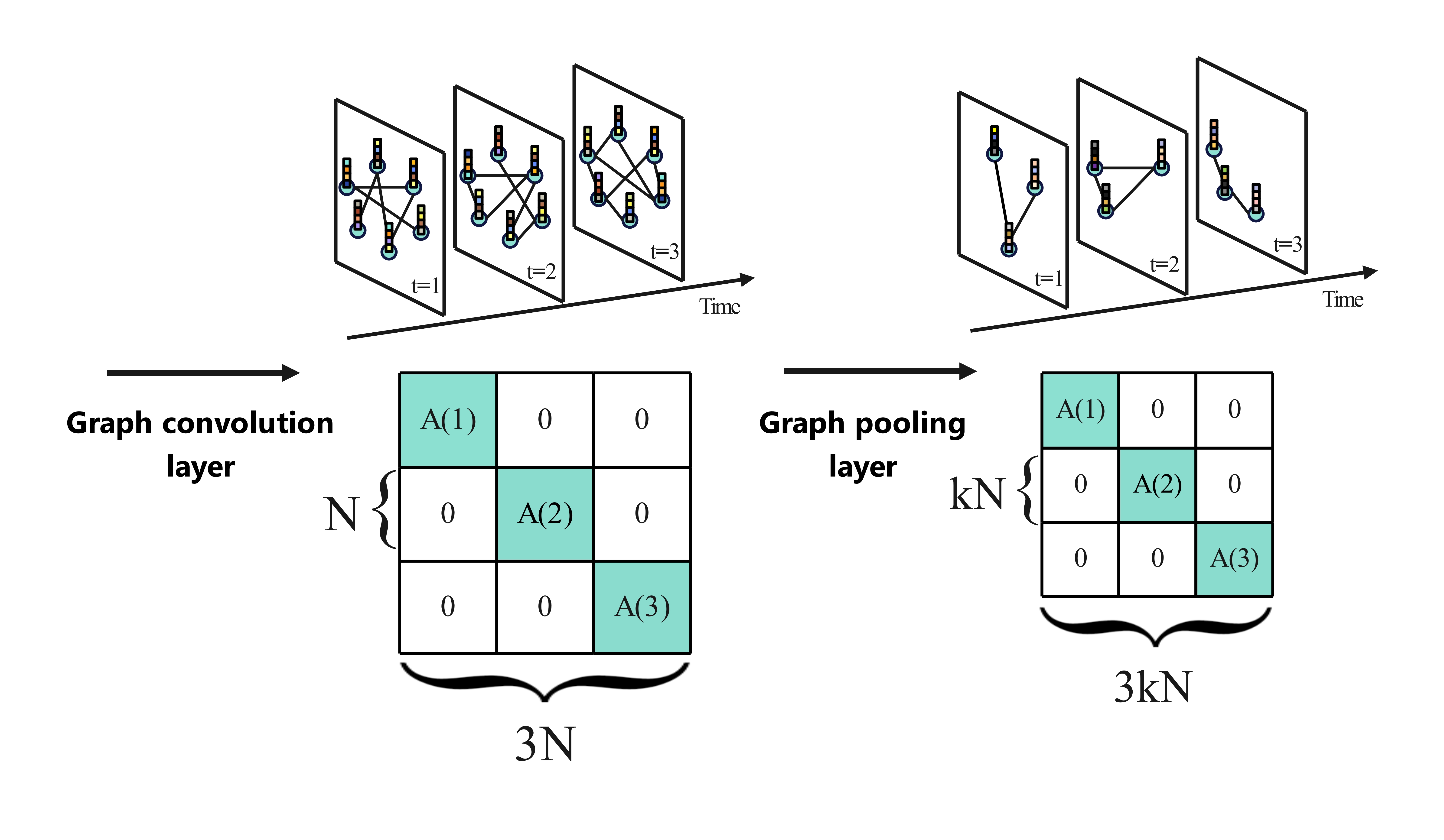}
     \caption{The dynamic brain graph and its associated adjacency matrix, before and after passing through the SAGPool layer, where $ N $ is the number of ROIs and $ k \in  [0, 1)$ indicates the pooling ratio.}\label{fig3}
   \end{minipage}
\end{figure}

\subsection{Hierarchical-graph based}
Recently, as shown in Fig \ref{fig0}(c), hierarchical graph-based methods have been introduced that leverage both brain network and population graph data. Jiang et al. \citep{JIANG2020104096}
 presented a hierarchical framework consisting of two levels. The first level generates embeddings by taking brain networks as input. The second level, considering these embeddings as node features, constructs a population graph by defining edges based on phenotype features. Then, they use a graph neural network to extract representations and perform classification. The entire process is trained end-to-end. Zhang et al. \citep{zhang2022classification} also employed a hierarchical framework, similar to the previous study, but with the difference that instead of using manual similarity metrics, they utilized a MLP to determine the similarity between samples. One of the challenges of this method is the lack of graph structure correction during training. To address this issue, Zaripova et al. \citep{zaripova2023graph} proposed a hierarchical model called Graph-in-Graph, which has the capability of learning the adjacency matrix of the population graph. 
 
  However, all these methods still face challenges, including not utilizing temporal analysis models for ROI-time series and assuming static FC. 
\section{Proposed Method}
\label{section3}

\subsection{HDGL model}
The proposed model, as shown in Fig\ref{fig1}, takes ROI-time series  as input and performs four stages of processing to extract  representations for the subjects and classify them.We assume that we have $ m $ subjects, each containing $ N $ ROIs. In the following, the four main parts of the model are introduced briefly.

1) Part (a): This part takes a time series of $ N $ ROIs as input. Then, it divides them into $ T $ segments using the sliding window technique and calculates FC within each segment. After thresholding, $ T $ adjacency matrices are obtained over time.

2) Part (b): In order to extract features for the brain graph nodes obtained from the previous part, a GRU model is employed. This model takes time series from the start time to the last time the window is applied for each graph as input and generates a vector representation as output.

3) Part (c): This part takes G(1), ..., G(T) as input and performs spatial analysis using GCN, SAGpool, and mean readout, and temporal analysis using a Transformer followed by sum readout. Ultimately,  a vector representation is obtained for each subject.

4) Part (d): Based on the embedding representation learned in part (c) and the similarity of phenotype features, a population graph is constructed. The nodes of this graph are then processed by multi-layer GNNs to learn new representations before finally being classified.

From another perspective, the proposed model can be considered hierarchical with two levels. At the first level, which includes parts (a), (b), and (c), the model constructs the brain network graphs and learns embeddings for them. At the second level, which includes part (d), it constructs the population graph and learns embeddings for its nodes. 

In the following, the four parts of the proposed model are introduced in detail.

\subsubsection{Part (a): Adjacency Matrix Construction}
By applying a predefined brain atlas, we compute the mean values of the BOLD signal within each ROI and then normalize them to zero mean and unit variance.
After that, the ROI-time series matrix $P\in R^{N\times T_{max}}$ is extracted, where $ N $ represents the number of brain regions, and $T_{max}  $ is the maximum length of the time series. As shown in Fig \ref{fig1}(a), we employed the sliding window technique to construct dynamic FC. The temporal window of length $ \Gamma $ and stride $ S $ is shifted over the time series to divide them into $ T $ segments, where $ T $ is computed as:
\begin{equation}
	T=\left\lfloor(T_{max}-\Gamma)/S\right\rfloor.
\end{equation}

Then, within each segment, Pearson correlation is calculated to construct the FC matrix. By considering a threshold on the FC values, a binary adjacency matrix $A \in \{0,1\} ^{N \times N} $ is obtained. Finally, for each subject $ i $, we have $ A_{i}=(A(1),...,A(T)) $. As depicted in Fig \ref{fig2}, the overall adjacency matrix for each subject is constructed without considering the edges between graphs in different time segments.

\subsubsection{Part (b): Node Feature Extraction}
Some previous works typically defined node features without considering any temporal changes. For instance, they used ROI-time series \citep{JIANG2020104096,zhang2022classification} or correlation correlation coefficients \citep{li2021braingnn,zhao2022dynamic} to define node features. To address this issue, we utilized the idea proposed in \citep{kim2021learning}. According to this definition, the node feature $ v $ at time $ t $ is defined as:
\begin{equation}
	x_{v}(t)=W[e_{v} || \eta(t)]
\end{equation}
where $ e_{v}\in \mathbb{R}^{N} $ is a one-hot vector where each element represents a specific ROI. $\eta(t) \in \mathbb{R}^{D}$ is the output of a GRU network. As shown in Fig \ref{fig1}(b),  to generate features for graph t, the GRU takes  ROI-time series from 0 to $t_{t}$, which $t_{t}$ is the endpoint of sliding-window. Then, the output vector is concatenated to  one-hot vector $e_{v}$, and a linear mapping is performed on it using learnable parameters, $W\in \mathbb{R}^{D \times (N+D)}$. It should be noted that other nodes in graph t use the same GRU output vector and differ only in the definition of their one-hot vectors.

\subsubsection{Part (c): Brain Network Analysis}
The brain graph $ G(t)=(V(t),E(t)) $ is obtained from segment t, where $ V(t)=\{x_{1}(t),...,x_{N}(t)\} $ is the set of its nodes. The set of edges is defined as: 
\begin{equation}
	E(t)=\{\{x_{i}(t),x_{j}(t)\}|j \in N_{(i)}, i \in \{1,...,N\}\}
\end{equation}
and $ N_{(i)} $ represents the neighbors of the $ i $-th node.

In part (c), as shown in Fig\ref{fig1}(c), our aim is to obtain a vector representation $ h_{G_{dyn}} $ in the output by receiving $ G_{dyn}=(G(1),...,G(T)) $ as a sequence of brain graphs obtained from $ T $ segments. These embeddings will be used as input to the part (d). Therefore, if part (c) learns a more informative representation for dynamic graphs, it can significantly impact the performance of part (d). Part (c) can be considered as a combination of two sections. The first section consists of several layers of GCN, pooling layers, and readout layers, which receives $ G_{dyn} $ as input and learns its spatial embeddings. Finally, produces sequence of representations $ (h_{G(1)},...,h_{G(T)}) $ in  output. The second section consists of a Transformer encoder that learns the temporal embeddings of the output from the previous section and, after aggregating them, generates the $ h_{G_{dyn}} $ in its output.

In the following, the two sections are introduced in detail.
\subsubsection*{\textbf{Graph convolution network, Pooling network and Readout layer:}}
In this subsection, the spatial embedding of the graphs constructed in the previous part is learned using GCNs\citep{kipf2016semi}. This network takes the adjacency matrix constructed in part (a), $ A \in \mathbb{R}^{N \times N} $, along with the feature matrix constructed in part (b), $ X \in \mathbb{R}^{N \times D}  $, and generates a new feature matrix as follows:
\begin{equation}
Z=\sigma(\tilde{D}^{-\frac{1}{2}}\tilde{A}\tilde{D}^{-\frac{1}{2}}X\theta)
\end{equation}
where $ \tilde{A}=A+I_{N} $, $I_{N}$ is an identity matrix and $ \tilde{D}\in \mathbb{R}^{N \times N}  $ is the degree matrix of $ \tilde{A} $, $ \sigma $ is ReLU activation function and $  \theta $ is a learnable matrix with a hidden dimension of $ D $.

Subsequently, a graph pooling layer is employed to find a simpler and more appropriate structure. In this model, the graph pooling layer, specifically the SAGPool\citep{lee2019self}, is employed for this purpose. SAGPool employs a self-attention mechanism to identify the more important nodes and distinguish them from the less important ones. To be more precise, in each layer, it considers the predefined hyperparameter (pooling ratio) $ k \in  [0, 1)$ and preserves $\lceil kN \rceil $ nodes from the initial graph while removing the rest. According to Fig \ref{fig3}, the pooling operator is applied separately to all the graphs within different segments. By removing some nodes, it alters the structure and features of these graphs. Additionally, in this figure, the changes in the dimensions of the dynamic graph adjacency matrix before and after the application of the pooling layer can be observed.

Using this approach can be beneficial from several perspectives. Firstly, the pooling layer reduces the size of the graphs, thus decreasing the number of model parameters and preventing overfitting. Secondly, by adjusting the graph structure, it has a positive impact on the performance of GNNs \citep{JIANG2020104096}.

Following this, the node representations are passed through a Readout layer, which aggregates them by taking their average to produce a single representation for each graph. Consequently, at the end of this stage, we obtain $ T $ vector representations for each subject, which will be further utilized for temporal embedding learning using a Transformer.

\subsubsection*{\textbf{Transformer encoder:}}
The output obtained from the previous section $ H=h_{G(1)},...,h_{G(T)} \in \mathbb{R}^ {T\times D}  $, represents a sequence of graph features over time. In order to learn the temporal embeddings of these sequences, a single-headed Transformer \citep{vaswani2017attention} has been employed. Due to the fact that the sole objective here is to find a suitable representation for the inputs and classify them, only the encoder part of the Transformer has been utilized. In this network, the input matrix $ H $ is multiplied by learnable matrices,  $W^Q$, $W^K$, and $W^V \in \mathbb{R}^{D\times d}$, to create the query, key, and value matrices, respectively:

\begin{equation}
	H\times W^Q=Q \in \mathbb{R}^ {T\times d},
\end{equation}
\begin{equation}
	H\times W^K=K \in \mathbb{R}^ {T\times d},
\end{equation}	
\begin{equation}	
	H\times W^V=V \in \mathbb{R}^ {T\times d}.
\end{equation}

Then, according to \ref{eq1}, Scaled Dot-product Attention is employed, resulting in the attention matrix. Intuitively, the model aims to preserve the values of the inputs it wants to focus on and discard irrelevant inputs by performing this operation.
\begin{equation}
\label{eq1}
	Attention(Q,K,V)=softmax(\frac{QK^T}{\sqrt{d_{k}}})V
\end{equation}
where $d_{k}$ is the dimension of the key matrix, denoted as $d$. 
In the following, the resulting matrix is passed through the feedforward neural network. After that, the sum of the computed output vector sequence is obtained and passed to a perceptron neural network. The resulting vector from this network is the final output $h_{G_{dyn}}$, which is then used as input to part (d).

\subsubsection*{\textbf{Multilayer design:}}
What was previously mentioned assumed that part (c) is single-layered. However, it can be designed as a multi-layered part.
According to Fig \ref{partc}, dynamic graphs, after passing through one GCN layer and one pooling layer, in the second layer, their representations are once again subjected to a GCN and pooling operation. Increasing the number of layers results in node representations being derived through deeper graph exploration, and subsequently, temporal relationships between these representations are learned by the Transformer. With the assumption of having $ K $ layers and considering $h^{(k)}_{G_{dyn}}$ as the representation of the graph in the $ k $-th layer, the final representation is computed according to:
\begin{equation}
	h_{G_{dyn}}=concatenate(\{h^{(k)}_{G_{dyn}}|k\in\{1,...,K\}\}).
\end{equation}

\subsubsection{Part (d): Population Network Analysis}
As previously described, the previous part takes a sequence of brain graphs over time, and by learning their spatial and temporal embeddings, it generates a vector representation for each subject in the output. In part (d), as shown in Fig\ref{fig1}(d), by receiving these vector representations and phenotypic information, 
we aim to first construct the population graph and then learn the embeddings of its nodes and perform classification. The following subsections provide an explanation of these steps:
\begin{figure}[!h]
	\begin{center}
		\includegraphics[scale=.38]{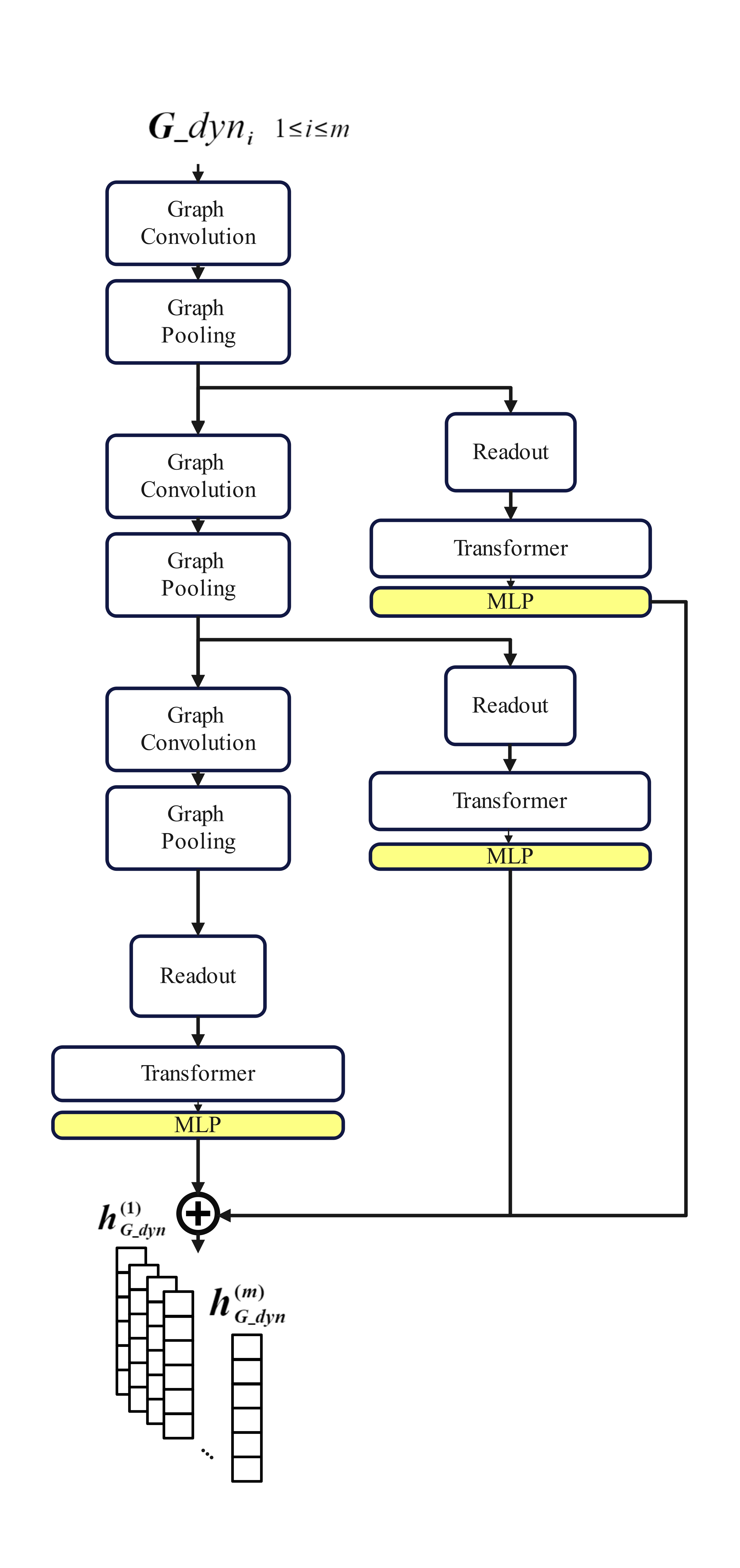}
		\caption{\label{partc}The general outline of part (c) for $ K $=3.}	
	\end{center}	
\end{figure}
\subsubsection*{\textbf{Constructing the Population Graph:}}
Creating the population graph requires two essential definitions: how to define the feature vectors of the nodes and how to consider the edges and edge weights. Nodes in the population graph represent various subjects. The feature considered for these nodes is a set of vector representations $\{h_{G_{dyn}}^{(1)},...,h_{G_{dyn}}^{(m)}\}$, which are the output of the part (c). In defining the edges, the assumption is made that phenotypic information, such as age and gender, can contain important insights into how different subjects (nodes) are related to each other. The aim is to construct a graph based on this information that can subsequently enhance the performance of the GNN. Therefore, the selection of which phenotypic information to use is crucial to represent the similarity between subjects and their imaging data as effectively as possible. Considering $ H $ as sets of phenotypic information (e.g., age, gender and site), the similarity between the phenotypic information of subjects, as denoted by $ S_{NI} $, is defined according to:
\begin{equation}
	\label{eq10}
	S_{NI}(v,w)=\sum_{h=1}^{H} \gamma(M_{h}(v),M_{h}(w)).
\end{equation}
In this equation, $ M_{h}(i) $ represents the $h$-th phenotype feature  of  the $i$-th subject, and $\gamma$ is a function for measuring the similarity between phenotypic features.  
\begin{figure*}[!h]
	\begin{center}
		\includegraphics[scale=.5]{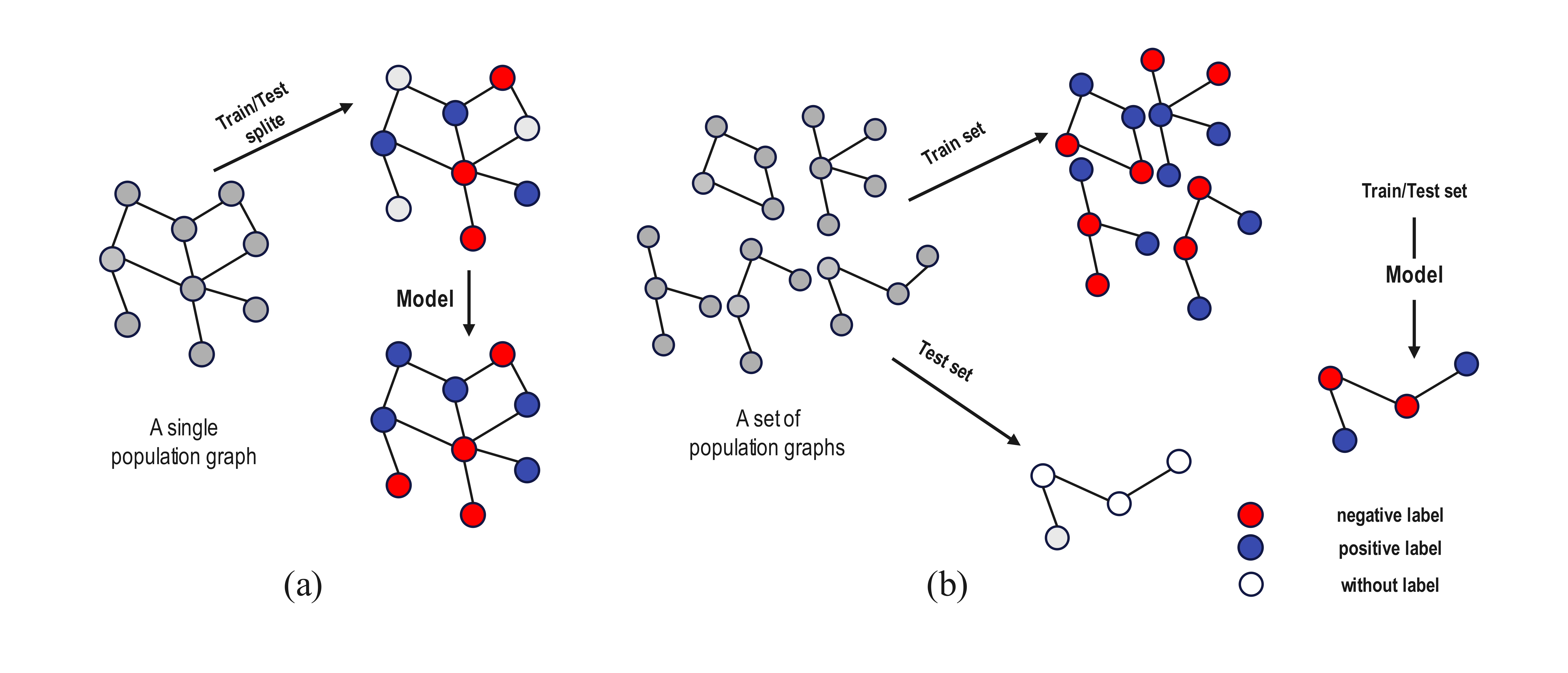}
		\caption{\label{fig4}(a) The process of forming a single population graph and separation of training and testing samples in transductive methods. (b) The procedure for constructing a set of population graphs and separation of training and testing samples in inductive method.}	
	\end{center}	
\end{figure*}
In \citep{parisot2017spectral}, it is stated that subjects belonging to the same class typically exhibit greater similarity in their brain graphs compared to subjects not belonging to the same class.  Therefore, imaging information is also utilized to obtain the edge weights. In the proposed model, we have employed this concept, and the similarity between the vector representations of subjects is utilized in the following manner:
\begin{equation}
	S_{I}(v,w)=exp(-\frac{[\rho(h_{G_{dyn}}^{(v)},h_{G_{dyn}}^{(w)})]^{2}}{2\sigma^2}).
\end{equation}
In this equation, $ h_{G_{dyn}}^{(i)}$ represents the representation obtained from part (c) for the $ i $-th subject. $ \rho$ denotes the correlation distance, and $\sigma$ represents the average values of $[\rho(h_{G_{dyn}}^{(v)},h_{G_{dyn}}^{(w)})]^{2}$.
In this way, we can utilize both image and non-image data to determine the similarity between subjects and the weights of edges:
\begin{equation}
	\label{eq12}
	S(v,w)=S_{I}(v,w)*S_{NI}(v,w).
\end{equation}

After completing the above steps, the population graph is generated, and subsequently, to find a suitable representation, it is fed into one of the GNN architectures that will be introduced in the next subsection.

\subsubsection*{\textbf{Graph Transformer:}}
In order to learn the population graph embedding, a type of GNN called a Graph Transformer \citep{shi2020masked} is used. Considering the node representation matrix in the $ l $-th layer as $ H^{(l)}=\{h^{(l)}_{1},h^{(l)}_{2},...,h^{(l)}_{m}\} $, where $ m $ represents the total number of subjects, similar to what was described in subsection 3.1.3 regarding the operation of Transformers, here, scaled dot-product attention is used to calculate attention coefficients. For each edge from node $ i $ to node $ j $, the attention coefficient is calculated by:
\begin{equation}
	\alpha_{i,j}=softmax(\frac{(W_{3}^{(l)}h_{i}^{(l)})^{T}(W_{4}^{(l)}h_{j}^{(l)})
	}{\sqrt{d_{k}}}).
\end{equation}
In this formula, $W_{3}^{(l)}$ and $ W_{4}^{(l)}\in \mathbb{R}^{d}$ represent learnable parameters, and $d_{k}=d $. Then, the node representation of node $i$ is updated as follows:
\begin{equation}
	h_{i}^{(l+1)}=W_{1}^{(l)}h_{i}^{(l)}+\sum_{j\in N(i)}\alpha_{i,j}W_{2}^{(l)}h_{j}^{(l)}.
\end{equation}
In this formula, $ N(i) $ represents the set of neighboring nodes of node $ i $, and $W_{1}^{(l)} $ and $ W_{2}^{(l)}\in \mathbb{R}^{d} $  are learnable parameters.

After the population graph passes through two Graph Transformer layers, the obtained representation for the nodes is passed through a linear layer for classification.
\begin{figure*}[!h]
	\begin{center}
		\includegraphics[scale=.5]{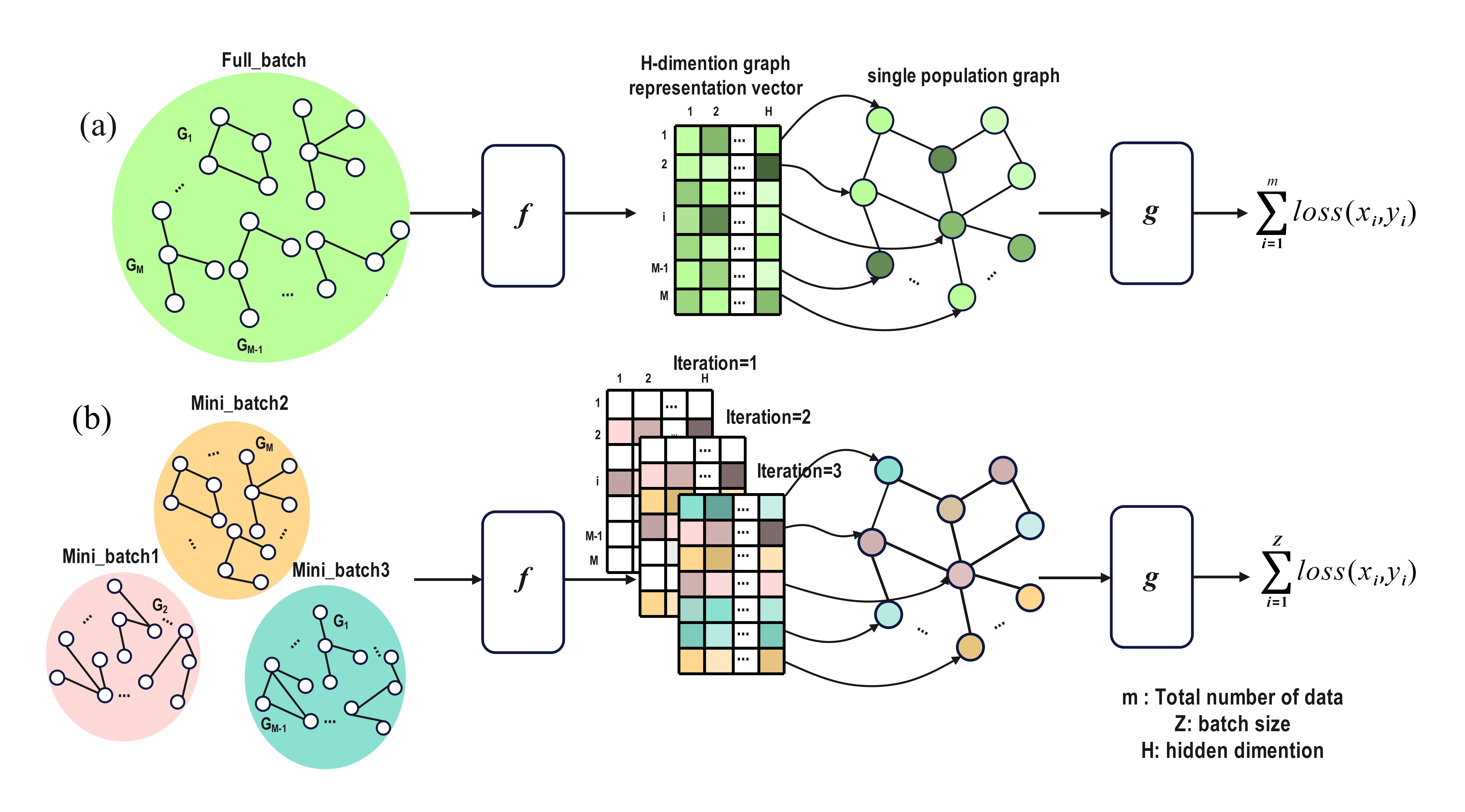}
		\caption{\label{fig5}The training procedure for the first and second levels of (a)HDGL$_{trans-join}$ and (b)	HDGL$
			_{trans-scl}$}	
	\end{center}	
\end{figure*}

\subsection{Training Methods}
As previously mentioned, the proposed model is a hierarchical model consisting of two levels. In the following, four different training methods are suggested based on how these two levels are trained.

\subsubsection{Transductive}
The methods in this category consistently use a fixed population graph with the same number of nodes as the total number of subjects. As shown in Fig\ref{fig4}(a), this population graph encompasses both training and test samples, allowing the model to benefit from test sample information during training. Training samples are depicted in blue and red based on their labels, while the testing samples are indicated in white. Consequently, these methods are considered semi-supervised. During training, loss is calculated solely based on the classification outcomes of the training data, while during testing, it relies exclusively on the classification results of the test data. In the following, two methods are proposed based on whether the first and second levels of the model are trained separately or end-to-end.

In the first method, the first and second levels are trained separately, each with its own specific loss function. The first level, after constructing brain graphs and obtaining their embeddings, utilizes a linear layer for classification. It learns based on the obtained loss.In the final epoch, the embeddings obtained are then provided as node features to the second level. At this level, first, a fixed population graph is constructed, and the embeddings are calculated. Then, classification is performed using a linear layer. Subsequently, the parameters of the second level are updated and trained. We have named the model trained using this method, HDGL$_{trans-sep}$. 

In the second method, the first and second levels are trained jointly in an  end-to-end fashion and use a common loss function. According to Fig\ref{fig5}(a), considering $ m $ as the total number of subjects, in each epoch, all subjects are fed into the first level, denoted by $ f $, and a vector representation with dimension $ H $ is generated as the output. These representations are then concatenated to form a matrix with dimensions $ m \times H $, which will serve as the feature matrix for the population graph at the second level, denoted by $ g $. After the formation of the population graph, the stages, as previously mentioned, are followed, and the subjects are classified, and the loss is calculated. We have named the model trained using this method, HDGL$_{trans-join}$.
In this scenario, the model must be trained in full-batch mode; therefore, if the number of training and testing data is high, it encounters the issue of memory shortage.

In order to address the high memory complexity issue in the second method, the third method attempts to mitigate memory complexity by considering smaller batch sizes while still forming a fixed population graph. In this method, samples are randomly divided into batches of size $ Z $, and one of these batches is provided as input to the first level at each iteration. The vector representation obtained for these $ Z $ samples is embedded in a matrix with dimensions of $ m \times H $, while the values of the representation for the remaining samples are assumed to be zero.
\begin{table*}[!h]
	\begin{center}
		
		\caption{\label{table1}Descriptions of the ABIDE and ADHD-200 datasets}	
		\begin{tabular}{lccccc}
			\hline
			& \multicolumn{1}{l}{Normal/Patient} & \multicolumn{1}{l}{Male/Female} & \multicolumn{1}{l}{Age} & \multicolumn{1}{l}{Timeseries length} & \multicolumn{1}{l}{\#Subjects} \\ \hline
			ABIDE    & 266/246                            & 427/85                          & 6-50                    & 176-246                               & 512                            \\ \hline
			ADHD-200 & 322/274                            & 396/200                         & 7-17                    & 119-232                               & 596                            \\ \hline
		\end{tabular}
	\end{center}
\end{table*}
 As depicted in Fig\ref{fig5}(b), in the first iteration, certain rows of the resulting matrix, corresponding to samples from the first batch, contain non-zero values and are highlighted with a color spectrum. The remaining rows, which correspond to samples not present in the first batch, have zero values and are marked in white. The resulting matrix will be considered as the population graph feature matrix. Then, the representations of all nodes and their corresponding classes are determined. However, the loss is calculated based only on the $ Z $ input samples, and the model parameters will be updated accordingly. In the subsequent iteration, another $ Z $  samples are provided as input to the first level, and the obtained representations for them replace in the previous feature matrix. Consequently, in the second iteration,$ 2\times Z $ rows of the feature matrix have non-zero values, and the algorithm continues in a similar manner. After one epoch, all rows of the matrix eventually acquire non-zero values.

We've named the model trained using this method HDGL$_{{trans-scl}}$, a scalable version of HDGL$_{{trans-join}}$, enabling it to effectively handle larger datasets.

\subsubsection{Inductive}
Unlike previous methods that had semi-supervised learning, this approach is trained in a supervised manner. In accordance with Fig \ref{fig4}(b), training and test samples are separated from each other, and the training samples are divided into batches of a specific size. A population graph is constructed only with the samples that are in the same batch. In this way, sets of population graphs are provided for model training. After training, population graph(s) are formed using the test data. As evident, this method, unlike the second method, requires less memory for training. We've denoted the model trained with this method as HDGL$ _{induc} $.

\section{Experiments and results}
\label{section4}
\subsection{Datasets}
In this study, we evaluated our proposed model on two challenging datasets for graph classification: the ABIDE (Autism Brain Imaging Data Exchange) and ADHD-200 (Attention Deficit Hyperactivity Disorder).

\textbf{ABIDE:}
Di Martino et al. \citep{di2014autism} gathered brain imaging data from international medical and university centers and formed the ABIDE dataset in 2014. This dataset contains 1112 subjects, which are collected from 17 acquisition sites. According to \citep{abraham2017deriving}, a subset of 871 samples with higher quality has been selected and preprocessed using the Configurable Pipeline for the Analysis of Connectomes (C-PAC) \citep{craddock2013towards}. Due to the use of the sliding window method, it is necessary to match the length of time series. Additionally, because the proposed model  HDGL$_{trans-join} $ faces memory constraints during execution, reducing the number of samples results in less memory demand. Therefore, we selected 512 individuals whose time series lengths fall within the range of 176-256. Table \ref{table1} reports the details of this dataset.

\textbf{ADHD:}
In 2012, Milham et al. \citep{adhd2012adhd} undertook the collection of data from 8 independent imaging sites. This data included brain images and clinical information from individuals with Attention-Deficit/Hyperactivity Disorder (ADHD) and healthy individuals. In this research, the preprocessed version of this dataset processed using the Athena pipeline \citep{bellec2017neuro}, was employed. In order to match the lengths of various time series and due to the incomplete personal characteristic
data (PCD), we retained only 596 subjects belonging to three sites: KKI, NYU, and Peking, while the rest were excluded. The details of this dataset are provided in \ref{table1}.

For both datasets, time series data of 116 different brain regions were extracted using the Automated Anatomical Labeling (AAL) atlas \citep{tzourio2002automated} and then normalized to have a zero mean and unit variance.
\subsection{Implementation details}
PyTorch and PyTorch Geometric libraries were used for implementing neural networks, and the nilearn library was employed to compute the FC of various subjects. The model was trained on a single GPU with 16GB of memory and 16GB of RAM. We used One-cycle Learning Rate Policy \citep{smith2019super} in which the learning rate starts from an initial value of 0.0005 and increases to its maximum after the first 20\% of training steps, reaching 0.0009, and then decreases. The length and stride of the sliding window are set to 20 and 5 timepoints (TPs), respectively. We also set the pooling ratio k = 0.8, embedding dimension D = d = 128, number of Transformer/Transformer encoder head = 1, hidden dimension of Transformer encoder = 16, number of layers K = 2 and 1, number of Transformer encoder layer = 2 and 3, dropout = 0.5 and 0.3 for ABIDE and ADHD-200, respectively. For models HDGL$ _{trans-scl} $, second-level of HDGL$ _{trans-sep} $, and HDGL$ _{induc} $, which do not fully utilize all batches, batch sizes of 120, 16, and 100 have been respectively considered. Also, we use the cross-entropy loss and the Adam optimizer.
\subsection{Statistical metrics}
In this research, we assess the network's performance by employing commonly used evaluation metrics, including Accuracy (ACC), Area Under Curve (AUC) and F1 score, to distinguish between brain disorder (positive) and healthy control (negative) samples. True Positive (TP) refers to the accurate classification of the positive class, whereas True Negative (TN) represents the correct classification of the negative class. On the other hand, False Positive (FP) signifies the incorrect prediction of the positive class, and False Negative (FN) indicates the inaccurate prediction of the negative class. The definitions of these evaluation metrics are as follows:
\begin{equation}
Accuracy=\frac{TP+TN}{TP+FP+TN+FN},
\end{equation}

\begin{equation}
Recall=\frac{TP}{TP+FN},
\end{equation}
\begin{equation}
Precision=\frac{TP}{TP+FP},
\end{equation}
\begin{equation}
F1-score=2*\frac{Precision*Recall}{Precision+Recall}.
\end{equation}

Also AUC is a metric that quantifies a classifier's ability to distinguish between positive and negative classes across different probability thresholds. A higher AUC indicates a better discriminative power of the model, with a perfect model having an AUC of 1 and a random model having an AUC of 0.5.

\begin{table*}[!h]
	\centering
	\caption{\label{table2}Performance comparison of various methods on ABIDE and ADHD-200 datasets, where the color of red and blue denote the top two best results, respectively.  }
	\begin{tabular}{llcccll}
		\hline
		Dataset                  & Model              & \multicolumn{1}{l}{\begin{tabular}[c]{@{}l@{}}Multi-\\ Modal\end{tabular}} & \begin{tabular}[c]{@{}c@{}}Type\\ FC\end{tabular} & ACC                                                  &   \multicolumn{1}{c}{F1}           & \multicolumn{1}{c}{AUC}          \\ \hline
		& SVM                & $\times$                                                                          & Static                                            & 61.5$\pm$5.21                                            &   \multicolumn{1}{c}{58.8$\pm$7.12}    & \multicolumn{1}{c}{62.4$\pm$5.45}    \\
		& Random forest      & $\times$                                                                          & Static                                            & 59.5$\pm$3.27                                            &   \multicolumn{1}{c}{59.1$\pm$2.54}    & \multicolumn{1}{c}{64.1$\pm$4.51}    \\
		& Decision tree      & $\times$                                                                          & Static                                            & 55.8$\pm$1.67                                            &   \multicolumn{1}{c}{53.6$\pm$2.81}    & \multicolumn{1}{c}{56.0$\pm$2.01}    \\
		& KNN                & $\times$                                                                          & Static                                            & 55.2$\pm$7.41                                            &   \multicolumn{1}{c}{47.9$\pm$5.30}    & 57.9$\pm$4.12                        \\
		& GCN                & $\times$                                                                          & Static                                            & 59.4$\pm$3.16                                            &   \multicolumn{1}{c}{59.5$\pm$3.08}    & 60.7$\pm$3.89                        \\
		& LSTM               & $\times$                                                                          & Static                                            & 61.2$\pm$1.99                                            &   \multicolumn{1}{c}{61.9$\pm$1.93}    & 64.9$\pm$2.33                        \\
		& Pop-GCN            &\checkmark                                                                           & Static                                            & 62.1$\pm$2.98                                            &   \multicolumn{1}{c}{64.6$\pm$2.45}    & 69.4$\pm$4.55                        \\
		& Hi-GCN             &\checkmark                                                                           & Static                                            & 65.5$\pm$3.18                                            &   \multicolumn{1}{c}{65.7$\pm$3.60}    & 73.0$\pm$4.23                        \\
		& STAGIN             & $\times$                                                                          & Dynamic                                           & 63.6$\pm$5.75                                            &   \multicolumn{1}{c}{63.2$\pm$5.70}    & 68.2$\pm$7.53                        \\
		& DGL                & $\times$                                                                          & Dynamic                                           & \multicolumn{1}{l}{65.3$\pm$3.74}                        &  65.1$\pm$2.88                        & 70.0$\pm$4.88                        \\
		& HDGL$_{ trans-sep} $    & \checkmark                                                                        & Dynamic                                           & \multicolumn{1}{l}{{\color[HTML]{3531FF} 70.3$\pm$3.69}} &  {\color[HTML]{3531FF} 70.2$\pm$3.73} & {\color[HTML]{3531FF} 75.1$\pm$3.64} \\
		& HDGL$_{ trans-join} $ & \checkmark                                                                          & \multicolumn{1}{l}{Dynamic}                       & \multicolumn{1}{l}{{\color[HTML]{FE0000} 72.4$\pm$4.79}} &  {\color[HTML]{FE0000} 72.3$\pm$5.01} & {\color[HTML]{FE0000} 76.5$\pm$3.03} \\
		& HDGL$_{ trans-scl} $    & \checkmark                                                                          & \multicolumn{1}{l}{Dynamic}                       & \multicolumn{1}{l}{66.6$\pm$3.01}                        &  64.7$\pm$3.24                        & 73.2$\pm$4.15                        \\ 
        
	\multirow{-14}{*}{ABIDE} & HDGL$_{ induc   } $ & \checkmark                                                                         & \multicolumn{1}{l}{Dynamic}                       & \multicolumn{1}{l}{66.4$\pm$3.77}                                            & 68.9$\pm$4.11                        & 71.3$\pm$5.20                        \\ \hline
		& SVM                & $\times$                                                                          & Static                                            & 57.4$\pm$4.17                                            &  51.2$\pm$6.01                        & 60.0$\pm$4.35                       \\
		& Random forest      & $\times$                                                                          & Static                                            & 54.7$\pm$3.80                                            &  48.6$\pm$4.64                       & 57.2$\pm$3.69                        \\
		& Decision tree      & $\times$                                                                          & Static                                            & 55.2$\pm$4.65                                            &  50.3$\pm$4.54                        & 54.6$\pm$4.32                        \\
		& KNN                & $\times$                                                                          & Static                                            & 54.5$\pm$2.27                                            &  55.2$\pm$1.85                        & 58.5$\pm$3.09                        \\
		& GCN                & $\times$                                                                          & Static                                            & 56.4$\pm$3.63                                            &  57.8$\pm$4.25                        & 57.3$\pm$5.16                        \\
		& LSTM               & $\times$                                                                          & Static                                            & 60.9$\pm$1.64                                            &  56.2$\pm$6.79                        & 61.8$\pm$3.37                        \\
		& Pop-GCN            &\checkmark                                                                         & Static                                            & 64.1$\pm$3.28                                            &  60.8$\pm$4.05                        & 67.3$\pm$3.52                        \\
		& Hi-GCN             & \checkmark                                                                        & Static                                            & 63.5$\pm$5.12                                            &  60.5$\pm$2.78                        & 64.7$\pm$3.09                        \\
		& STAGIN             & $\times$                                                                          & \multicolumn{1}{l}{Dynamic}                       & 62.4$\pm$4.63                                            &  59.4$\pm$4.98                        & 64.6$\pm$5.57                        \\
		& DGL                & $\times$                                                                          & \multicolumn{1}{l}{Dynamic}                       & \multicolumn{1}{l}{62.9$\pm$5.82}                        &  61.7$\pm$4.84                        & 66.6$\pm$4.98                        \\
		& HDGL$_{trans-sep}  $ & \checkmark                                                                         & \multicolumn{1}{l}{Dynamic}                       & \multicolumn{1}{l}{{\color[HTML]{3531FF} 65.3$\pm$5.43}} &  {\color[HTML]{3531FF} 64.0$\pm$2.34} & {\color[HTML]{3531FF} 69.1$\pm$3.33} \\
		& HDGL$ _{trans-join} $ &\checkmark                                                                          & \multicolumn{1}{l}{Dynamic}                       & \multicolumn{1}{l}{{\color[HTML]{FE0000} 65.7$\pm$3.67}} &  {\color[HTML]{FE0000} 64.6$\pm$1.87} & {\color[HTML]{FE0000} 70.4$\pm$4.21} \\
		& HDGL$ _{trans-scl} $    &\checkmark                                                                          & \multicolumn{1}{l}{Dynamic}                       & \multicolumn{1}{l}{64.3$\pm$4.09}                        &  63.0$\pm$5.90                        & 68.8$\pm$5.03                        \\
		\multirow{-14}{*}{ADHD-200}  & HDGL$ _{induc} $   &\checkmark                                                                          & \multicolumn{1}{l}{Dynamic}                       & \multicolumn{1}{l}{65.0$\pm$5.32}                        &  63.1$\pm$3.76                        & 68.7$\pm$4.78                        \\ \hline
	\end{tabular}
\end{table*}

\subsection{Comparison with the baselines}
We compare the proposed HDGL model and its variant with both deep learning-based methods and machine learning methods include 
\textbf{SVM} \citep{vapnik1999nature}, \textbf{random forest} \citep{ho1995random}, \textbf{decision tree} \citep{belson1959matching} and \textbf{KNN} \citep{fix1985discriminatory}, which
were all implemented using the scikit-learn library\citep{pedregosa2011scikit}. These models can be categorized into two groups: unimodal models that solely utilize fMRI information for classification, and multimodal models that can also incorporate phenotypic information. Additionally, based on the type of FC they utilize, these models can be further divided into static and dynamic models. For  machine learning models, the upper triangle part of the FC matrix are considered as input features. In the following sections, the competing deep learning-based models are described.

\textbf{GCN} \citep{kipf2016semi}: First, a brain network is constructed by considering static FC as the adjacency matrix and using the ROI-time series as node features. After passing through a two-layer GCN, the node representations are averaged and classified with a linear layer.

\textbf{LSTM}: The input sequence of this model is represented as
$\textit{X}$=$x^1$,$x^2$,...,$x^T$, where at each time step $ t $, a vector of dimension $ N $ is fed into the model. Here, $ N $ represents the number of ROIs. Then, this input sequence is passed through a four-layer LSTM network and fed into a linear layer for classification with the sigmoid activation function.

\textbf{Pop-GCN} \citep{parisot2017spectral}: It is considered as population-based graph models and leverages both image and non-image information. The graph node features are determined through the use of RFE.

\textbf{Hi-GCN} \citep{JIANG2020104096}: The introduced method follows a hierarchical structure. Initially, brain networks are formed by utilizing FC and ROI-time series,and the embeddings are obtained using a GCN. Then, population graphs are constructed based on phenotype features, and classification is performed by passing them through multiple layers of the GCN.

\textbf{STAGIN} \citep{kim2021learning}: This model extracts FC
matrices using a sliding window over the ROI-time series, thus falls into the category of dynamic FC-based models. Additionally, as it only utilizes imaging data, it can be considered as a unimodal model.
\begin{table*}[!h]
	\begin{center}
		\caption{\label{table3}The average performance of the proposed model with the removal of its components on the ABIDE.}	
		\begin{tabular}{ccclllccc}
			\hline
			\multicolumn{9}{c}{ABIDE}                                                                                                                                                                                                                                               \\ \hline
			\multicolumn{1}{l}{GCN} & \multicolumn{1}{l}{GRU} & \multicolumn{1}{l}{SAGpool} & Transformer           & Pop-graph             &  & \multicolumn{1}{l}{ACC}&  \multicolumn{1}{l}{F1}                       & \multicolumn{1}{l}{AUC} \\ \hline
			\checkmark                       & \multicolumn{1}{l}{}    & \multicolumn{1}{l}{}        &                       &                       &  & 58.7                    &  \multicolumn{1}{l}{57.4} & 59.2                    \\
			*\checkmark                       & \checkmark                       &                             &                       &                       &  & 61.2                     & 59.9                     & 61.7                    \\
			*\checkmark                       & \checkmark                       & \checkmark                           &                       &                       &  & 62.4                     & 62.2                     & 64.8                    \\
			*\checkmark                       & \checkmark                       & \checkmark                           & \multicolumn{1}{c}{\checkmark} &                       &  & 65.3                   & 65.1                     & 70.0                    \\
			\checkmark                       & \checkmark                       & \checkmark                           & \multicolumn{1}{c}{\checkmark} & \multicolumn{1}{c}{\checkmark} &  & 72.4                   & 72.3                     & 76.5                    \\ \hline
		\end{tabular}
	\end{center}
\end{table*}

\begin{table*}[!h]
	\begin{center}
		\caption{\label{table4}The average performance of the proposed model with the removal of its components on the ADHD-200.}	
		\begin{tabular}{ccclllccc}
			\hline
			\multicolumn{9}{c}{ADHD-200}                                                                                                                                                                                                                                          \\ \hline
			\multicolumn{1}{l}{GCN} & \multicolumn{1}{l}{GRU} & \multicolumn{1}{l}{SAGpool} & Transformer           & Pop-graph            &  & \multicolumn{1}{l}{ACC} &  \multicolumn{1}{l}{F1} & \multicolumn{1}{l}{AUC} \\\hline
			\checkmark                       & \multicolumn{1}{l}{}    & \multicolumn{1}{l}{}        &                       &                       &  & 57.6                    &  59.3 & 58.1                    \\
			*\checkmark                       & \checkmark                       &                             &                       &                       &  & 58.7                    &   59.1                     & 61.4                    \\
			*\checkmark                       & \checkmark                       & \checkmark                           &                       &                       &  & 59.6                     & 59.2                     & 62.1                   \\
			*\checkmark                       & \checkmark                       & \checkmark                           & \multicolumn{1}{c}{\checkmark} &                       &  & 62.9                      & 61.7                    & 66.6                    \\
			\checkmark                       & \checkmark                       & \checkmark                           & \multicolumn{1}{c}{\checkmark} & \multicolumn{1}{c}{\checkmark} &  & 65.7                  & 64.6                     & 70.4                    \\ \hline
		\end{tabular}
	\end{center}
\end{table*}
\subsection{Classification Results}
We compared various proposed models of HDGL with competing models on the ABIDE and ADHD-200 datasets to evaluate their classification performance. We employed the stratified 5-fold cross-validation method to partition the training and test data, ensuring a fair comparison by maintaining consistent training and test sets for all models. We also evaluated the first-level performance of HDGL by adding a linear classifier to the end of part (c) and referred to it as DGL. The phenotypic information of the multimodal models includes age and gender for the ABIDE dataset and gender, imaging site, and age for the ADHD-200 dataset. The numbers reported in Table \ref{table2} are the average values of the 5 folds along with their standard deviations.

Based on the results, it can be observed that:
1) In general, deep learning-based models outperform classical machine learning methods. This issue may arise from the inherent complexity of fMRI data, where manual feature extraction often fails to yield satisfactory results.
2) The superior performance of LSTM compared to machine learning-based models and GCN may indicate the significance of temporal analysis of ROI-time series and the presence of influential patterns within them.
3) In general, models that operate based on dynamic FC exhibit better performance compared to models that operate based on static FC. For instance, the DGL model, which is based on dynamic FC, despite utilizing only imaging data, improves the accuracy metric by approximately 3.2 compared to the Pop-GCN model that leverages both imaging and non-imaging information. This may indicate that considering statistical dependencies between ROI-time series  in smaller time segments can contain important information about the dynamics of brain functioning over time.
4) After adding the population graph to DGL and training the multimodal models based on training methods one to four, it is observed that the AUC metric in the ABIDE dataset increases by 5.1, 6.5, 3.2, and 1.2, respectively, and in the ADHD-200 dataset, it increases by 2.5, 3.8, 2.2, and 2.1, respectively. This indicates the better performance of multimodal methods compared to unimodal methods and the impact of phenotype data on classification improvement. Additionally, this can indicate the impact of parameters such as age and gender on susceptibility to brain disorders.
5) The increase of 2.1 and 0.4 in accuracy for the HDGL$_{ trans-join} $ model compared to HDGL$_{trans-sep} $ indicates that end-to-end training of the model, as opposed to separate training of the first and second levels, results in the extraction of more useful features for reducing classification errors.
6) HDGL$_{trans-join} $ employs semi-supervised learning and utilizes test data information during training. In contrast, HDGL$ _{induc} $ is only allowed to use training data during its training process. This can justify the superior performance of the HDGL$ _{trans-join} $ compared to HDGL$_{induc}$.
7) The proposed model, HDGL$_{trans-join} $, outperforms the best multimodal model, Hi-GCN, with an increase of 6.9, 3.5, and 6.6 in accuracy, AUC, and F1, respectively. It also surpasses the best dynamic FC-based model, STAGIN, with an increase of 8.8, 8.3, and 9.1 in accuracy, AUC, and F1, respectively, demonstrating superior performance on the ABIDE dataset. On the ADHD-200 dataset, this model also outperforms Pop-GCN, with an increase of 1.6, 3.1, and 3.8 in accuracy, AUC, and F1, respectively. Additionally, it surpasses STAGIN, with an increase of 3.3, 5.8, and 5.2 in accuracy, AUC, and F1, respectively, showcasing its superior performance.

\section{Ablation study and discussion}
\label{section5}
\subsection{Assessing the efficiency of HDGL components}
To assess the effectiveness of model components in improving its overall performance, we conducted this experiment by sequentially adding components to the model and training it each time using stratified 5-fold cross-validation. The average model performance for the ABIDE and ADHD-200 datasets is reported in Tables \ref{table3} and \ref{table4}, respectively. It is worth mentioning that in the first experiment, where the model consists only of GCN, the time series in each segment are treated as the features of the nodes of the brain graph. Finally, the embeddings are averaged for each graph. Subsequently, these embeddings are aggregated to produce a representation for each subject, followed by classification. From the tables, it can be observed that adding each component has led to an improvement in model performance. Particularly, the population graph and Transformer have had the most significant impact on increasing efficiency. This highlights the importance of modeling the inter-subject relationships, utilizing phenotypic features, and considering the dynamic statistical dependencies among ROI-time series in classification.
\begin{figure}
	\centering
	
	\begin{subfigure}{0.48\textwidth}
		\includegraphics[width=\textwidth]{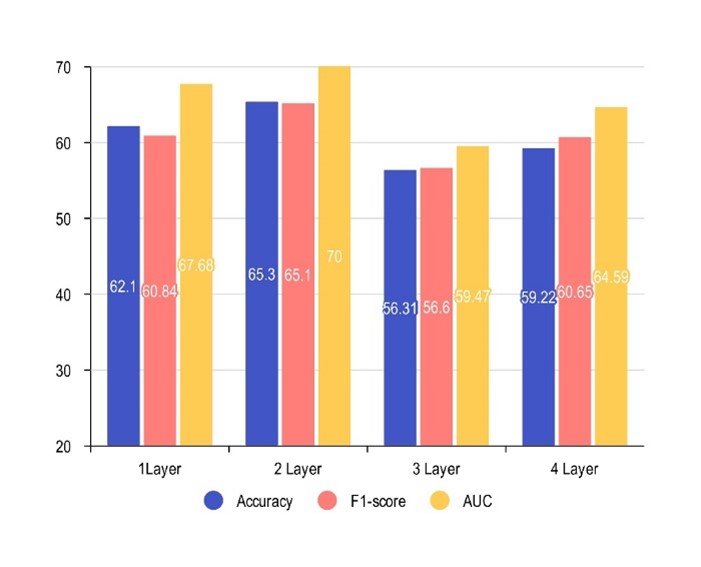}
		\caption{}
		\label{fig:first1}
	\end{subfigure}
	\hfill
	\begin{subfigure}{0.48\textwidth}
		\includegraphics[width=\textwidth]{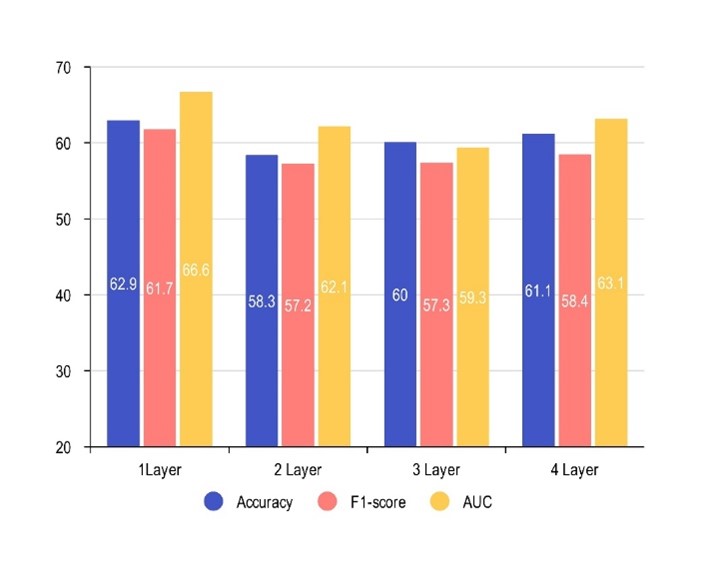}
		\caption{}
		\label{fig:second1}
	\end{subfigure}
	
	\caption{\label{fig6}The average accuracy, F1 score, and AUC of the proposed model at the first level for different number of layers on (a) ABIDE and (b) ADHD-200.}	
\end{figure}

\subsection{Number of layers}
As mentioned earlier, part (c) can be designed as a multi-layer. In this experiment, we varied the number of layers from one to four, and each time, we trained the first level using stratified 5-fold cross-validation. The average accuracy, F1 score, and AUC for both the ABIDE and ADHD-200 datasets can be observed in Fig\ref{fig6}(a) and \ref{fig6}(b), respectively. Based on the results, for the ABIDE dataset, increasing the number of layers up to two improves the model's performance, and after that, it starts to degrade. However, for the ADHD dataset, increasing the number of layers results in a performance drop, and the model achieves its best performance with one layer. Accordingly, considering second-order neighborhood information for the graphs in the ABIDE dataset leads to a better representation for each node.

\subsection{Scalability analysis of the HDGL \texorpdfstring{$ _{trans-scl} $}{Lg}}
As previously mentioned, HDGL$ _{trans-join} $ receives data in full batches, which led to memory limitations during execution. Consequently, we had to exclude some data to make this model implementable and executable. On the other hand, HDGL$ _{trans-scl} $ was introduced to reduce memory requirements at runtime. Unlike the previous model, this model, without memory constraints, can accept any number of data instances as input and is referred to as a scalable model. To validate this, in this experiment, we added 123 subjects from ABIDE, including 74 healthy and 49 ASD samples, to the training set. Both HDGL$ _{trans-join} $ and HDGL$ _{trans-scl}$  were trained using a  stratified 5-fold cross-validation, with the difference that 123 subjects were added to the HDGL$ _{trans-scl}$ training set, while the test sets were the same for both models in each fold.
 The average performance results of both models are reported in Fig\ref{fig7}. The results indicate the superiority of  HDGL$ _{trans-scl}$'s performance over HDGL$ _{trans-join} $ in this scenario. This indicates that the HDGL$ _{trans-scl}$, due to its lack of memory constraints during execution, is capable of achieving better performance than HDGL$ _{trans-join} $ by receiving more data during training.

\subsection{Impact of pooling ratio}
\begin{figure}[!h]
   \begin{minipage}{0.48\textwidth}
     \centering
     \includegraphics[width=1.155\linewidth]{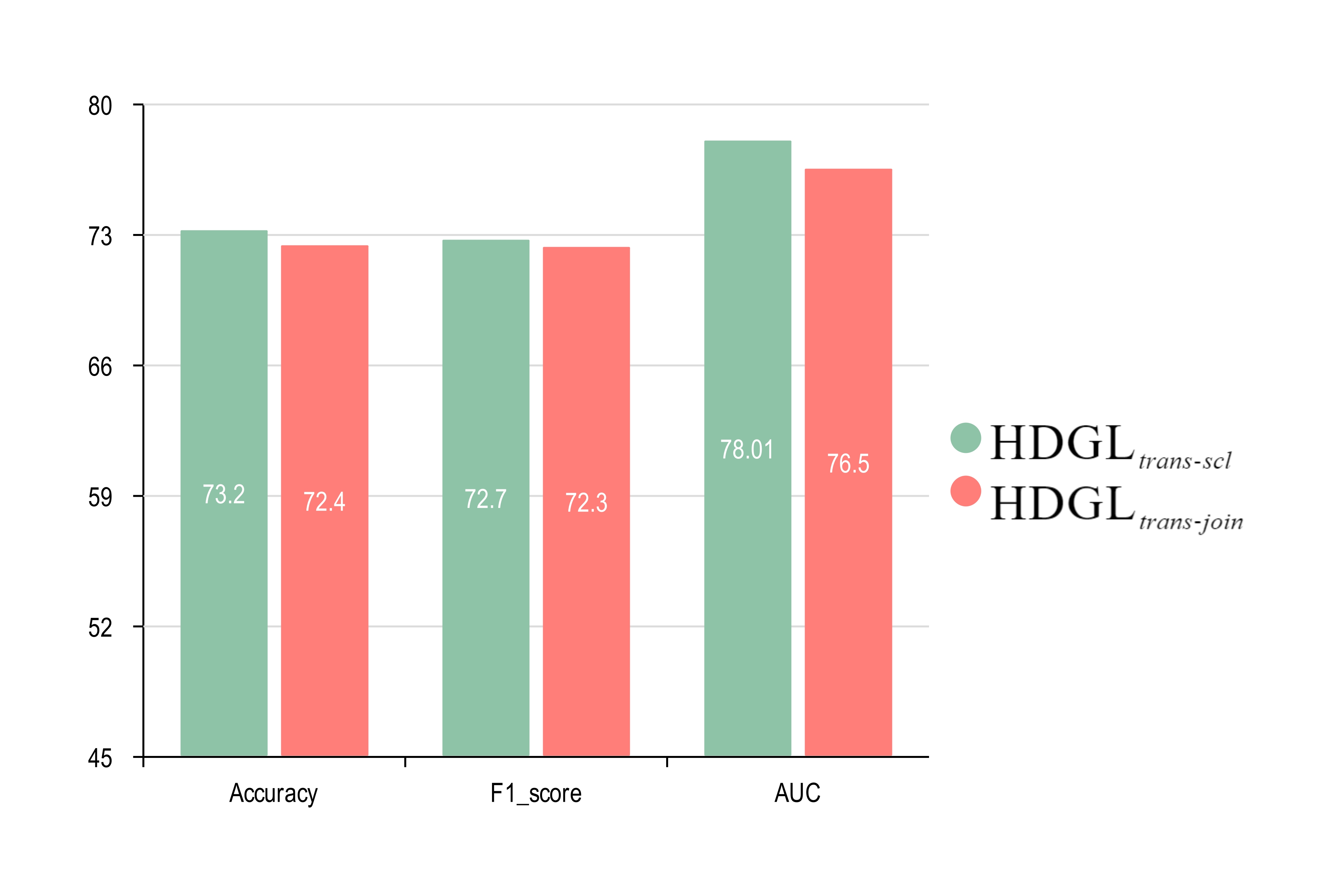}
     \caption{The average performance of the HDGL$ _{trans-join} $ (red) and the HDGL$ _{trans-scl} $ (green) with the addition of 123 samples to its training set.}\label{fig7}
   \end{minipage}\hfill
   \begin{minipage}{0.48\textwidth}
     \centering
     \includegraphics[width=1\linewidth]{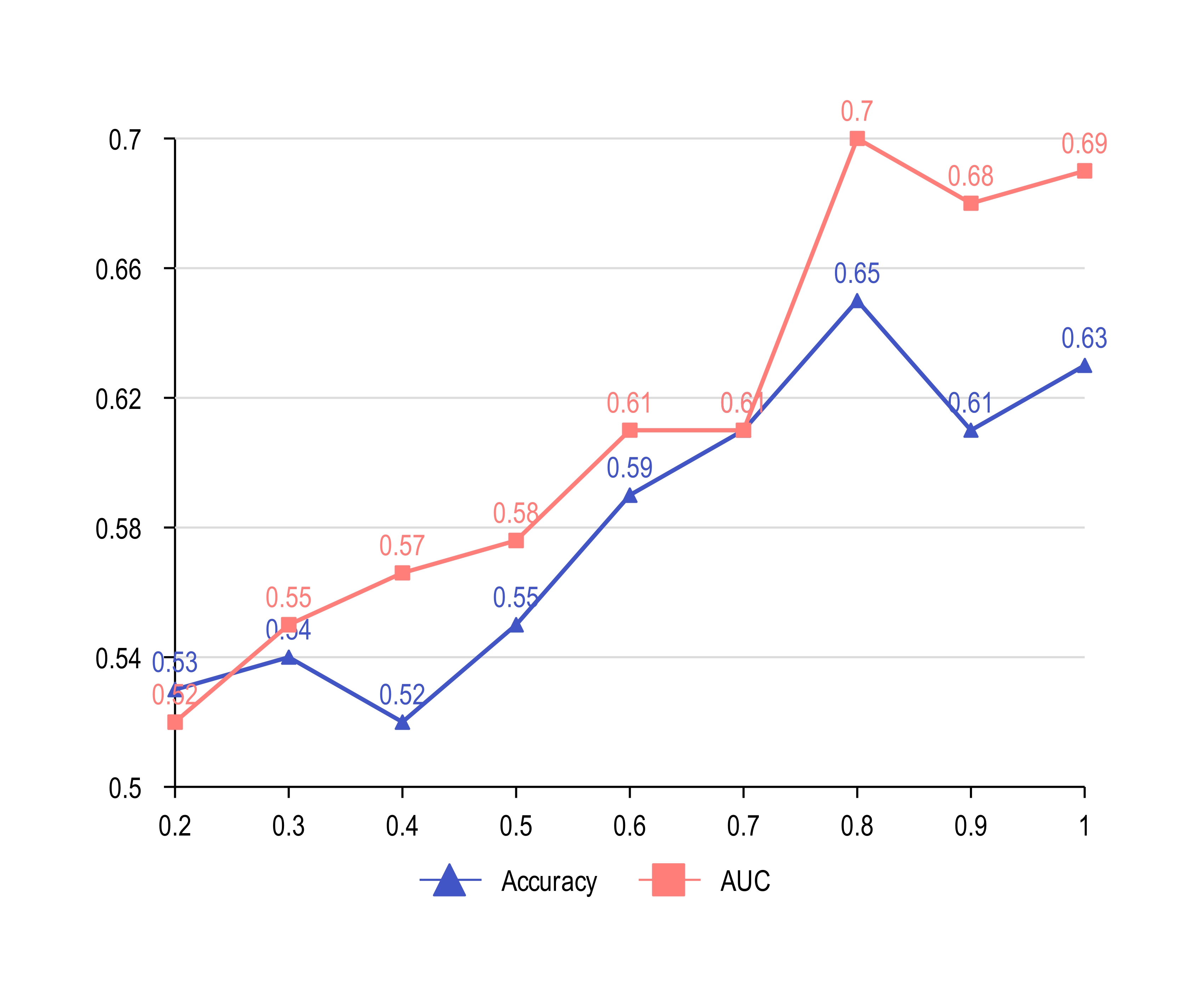}
     \caption{The changes in accuracy and AUC of the first-level of the proposed model for different pooling ratios.}\label{fig8}
   \end{minipage}
\end{figure}

Since the graph structure significantly affects the model's performance, the pooling ratio can be considered an important factor.
Considering very small values for it will force the model to eliminate some of the important and useful nodes and connections. On the other hand, considering a very high values will result in noisy  nodes and connections remaining intact, weakening the classifier's performance and leading to overfitting. In this experiment, we investigated the impact of different pooling ratios on the performance of the proposed model at the first level. Accordingly, the model was trained with a stratified 5-fold cross-validation with pooling ratios ranging from 0.2 to 1.0. The average accuracy and AUC are reported in Fig \ref{fig8}. With an increase in the pooling ratio, the model's performance improved overall, up to a pooling ratio of 0.8. Beyond this point, increasing the pooling ratio led to a decline in the model's performance.

\subsection{Influence of phenotypic information}
\begin{figure*}[!h]
	
		\centering
	\includegraphics[scale=0.6]{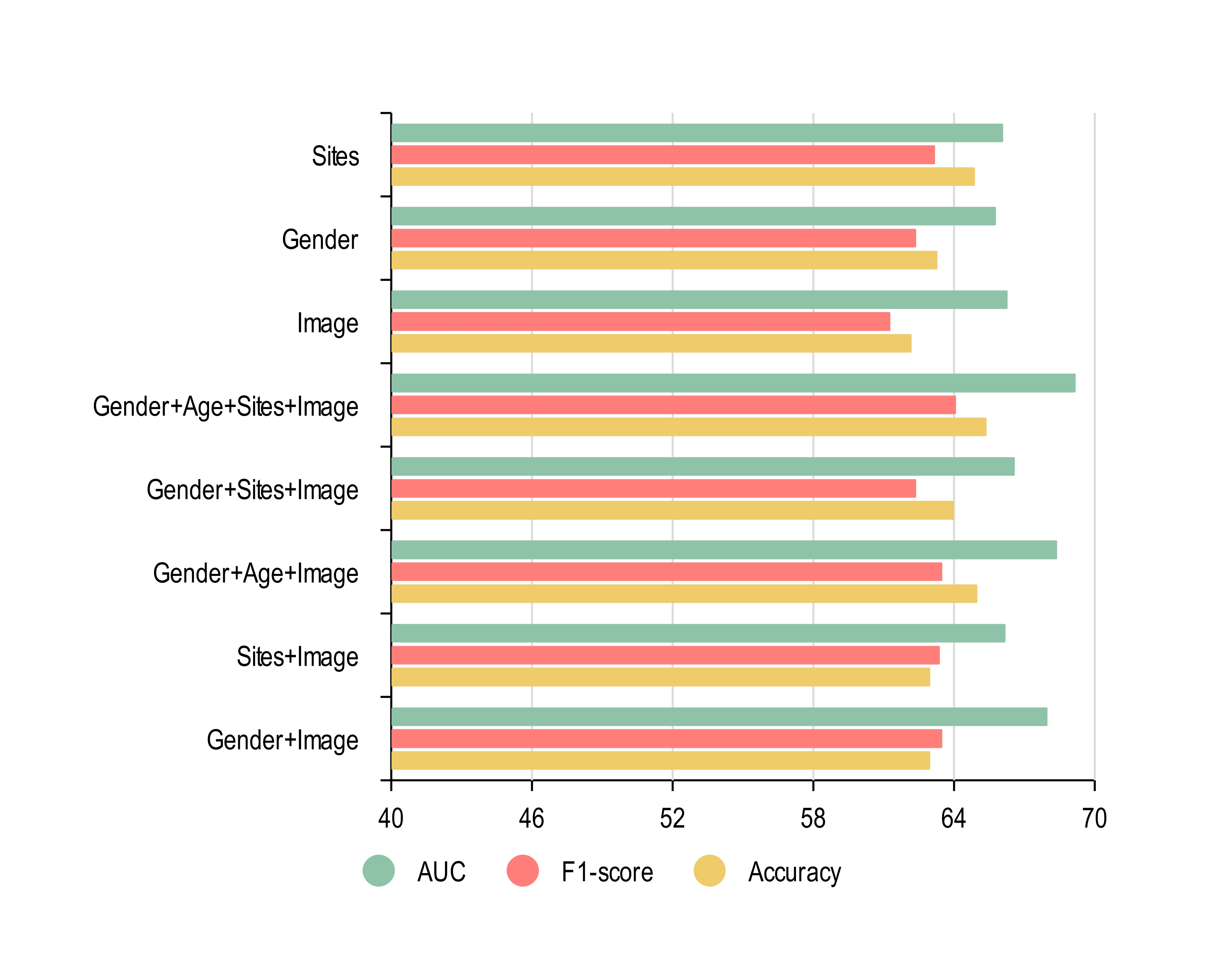}
	\caption{\label{fig9}The average evaluation results of the proposed model based on various combinations of phenotype features on the ADHD-200}	
\end{figure*}

The graph structure has a significant impact on the performance of GNNs. Therefore, in this experiment, we investigated the influence of various phenotypic features in constructing the population graph and, consequently, the performance of the proposed model. Subsequently, various scenarios were examined in which the edge weights of the population graph were determined based on non-imaging features such as age, gender, imaging site, and various combinations thereof, as described in Equation \ref{eq10}. Following this, with the addition of imaging features, the edge weights were once again computed according to different combinations of imaging and non-imaging features, as specified in Equation \ref{eq12}. The proposed model was trained in each experiment using stratified 5-fold cross-validation, and its hyperparameters were tuned. Subsequently, the average performance of the model on the ABIDE  is reported in Table \ref{table5}, and on the ADHD-200 in Fig \ref{fig9}. Based on the results, it can be seen that adding image features to phenotype features had a significant impact on improving the model's performance. For instance, according to Table \ref{table5}, considering only the site as a feature resulted in an average AUC of 69.9. In another experiment, with the addition of image features to the site, the average AUC increased to 73.7. Similarly, considering only the image feature for constructing the population graph resulted in an average AUC of 69.3, and this value consistently increased with the addition of other phenotype features.Ultimately, for ABIDE, the best performance is achieved through the combination of gender, age, and imaging. This finding aligns with previous research indicating the influence of gender on the likelihood of autism spectrum disorder \citep{zhang2022classification}. Additionally, for ADHD-200, the best performance is obtained with the combination of gender, age, site, and imaging.

\begin{table*}[!b]
\centering
	\fontsize{8pt}{10pt}\selectfont	
	\caption{\label{table5}The mean and standard deviation of the results of the proposed model based on different combinations of phenotypic features on ABIDE.}
	\begin{tabular}{llll}
		\hline
		Phenotypic information      & \multicolumn{1}{c}{Accuracy} & \multicolumn{1}{c}{AUC} & \multicolumn{1}{c}{F1-score} \\ \hline
		Sites                       & 62.7$\pm$5.64                    & 69.9$\pm$5.76               & 63.8$\pm$4.22                    \\
		Gender                      & 62.7$\pm$5.79                    & 69.8$\pm$4.85               & 63.0$\pm$5.63                    \\
		Image                  & 62.6$\pm$4.90                    & 69.3$\pm$3.40               & 64.6$\pm$5.73                    \\
		Sites + Gender              & 63.4$\pm$5.02                    & 71.1$\pm$6.18               & 64.4$\pm$3.01                    \\
		Gender + Age                & 62.4$\pm$2.36                    & 69.7$\pm$3.08               & 64.8$\pm$3.47                    \\
		Gender + Image         & 61.8$\pm$1.87                    & 70.6$\pm$2.58               & 63.3$\pm$3.97                    \\
		Sites + Image           & 69.6$\pm$3.74                    & 73.7$\pm$4.78               & 69.2$\pm$3.59                    \\
		Sites + Gender + Age        & 62.8$\pm$4.10                    & 69.8$\pm$4.35               & 64.2$\pm$5.02                    \\
		Gender + Age + Image   & \textbf{70.3$\pm$3.69}           & \textbf{75.1$\pm$3.64}      & \textbf{70.2$\pm$3.73}           \\
		Gender + Sites + Image & 68.1$\pm$4.28                    & 73.5$\pm$4.60               & 69.1$\pm$5.46                    \\ \hline
	\end{tabular}
\end{table*}
\begin{figure*}
\centering
\begin{subfigure}{0.297\textwidth}
		\includegraphics[width=\textwidth]{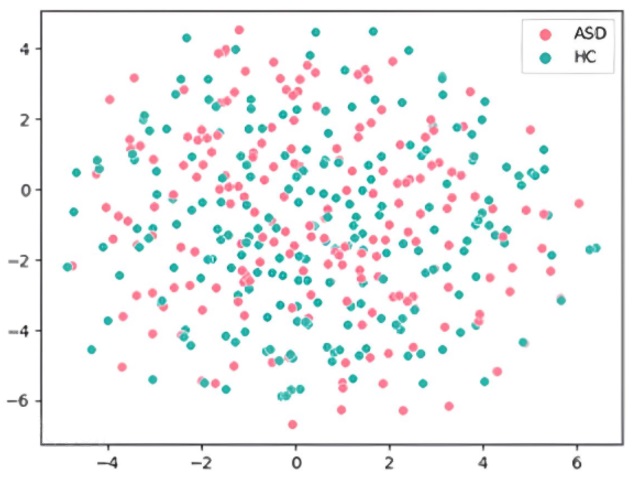}
		\caption{}
		\label{fig:first}
\end{subfigure}
\hfill
\begin{subfigure}{0.3\textwidth}
		\includegraphics[width=\textwidth]{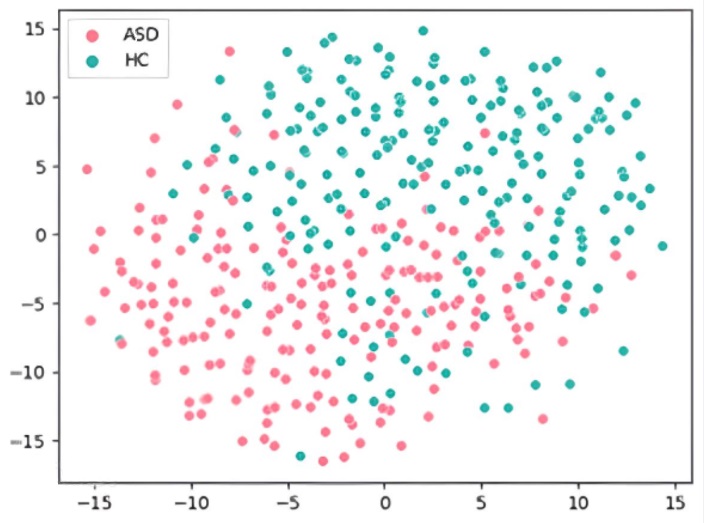}
		\caption{}
		\label{fig:second}
\end{subfigure}
\hfill
\begin{subfigure}{0.3\textwidth}
		\includegraphics[width=\textwidth]{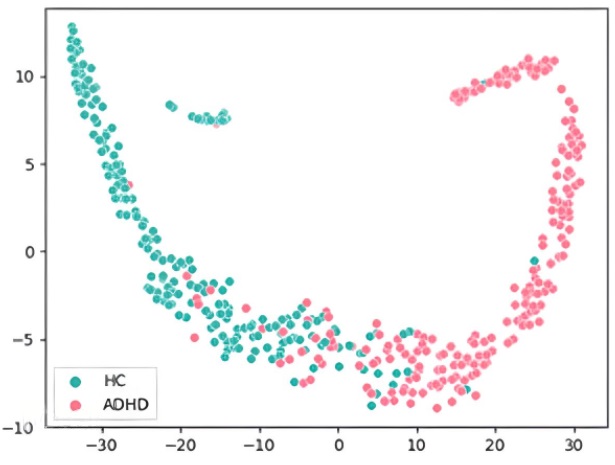}
		\caption{}
		\label{fig:third}
\end{subfigure}
\vfill
\begin{subfigure}{0.3\textwidth}
		\includegraphics[width=\textwidth]{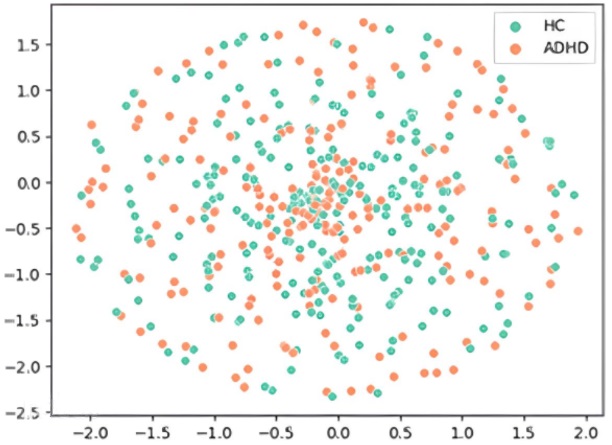}
		\caption{}
		\label{fig:forth}
\end{subfigure}
\hfill
\begin{subfigure}{0.3\textwidth}
		\includegraphics[width=\textwidth]{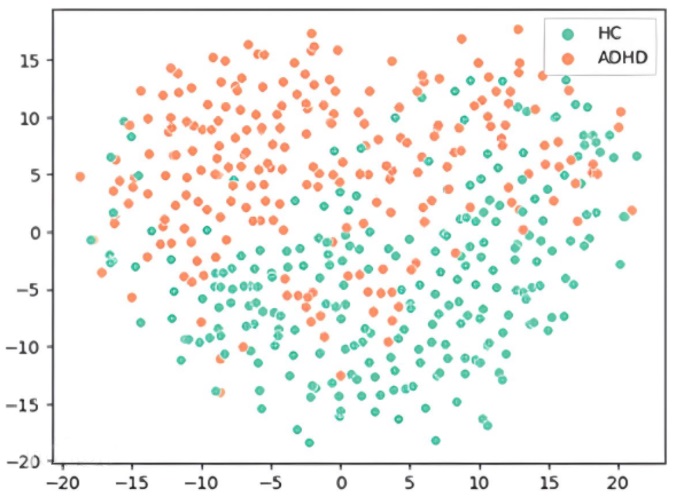}
		\caption{}
		\label{fig:fifth}
\end{subfigure}
\hfill
\begin{subfigure}{0.3\textwidth}
		\includegraphics[width=\textwidth]{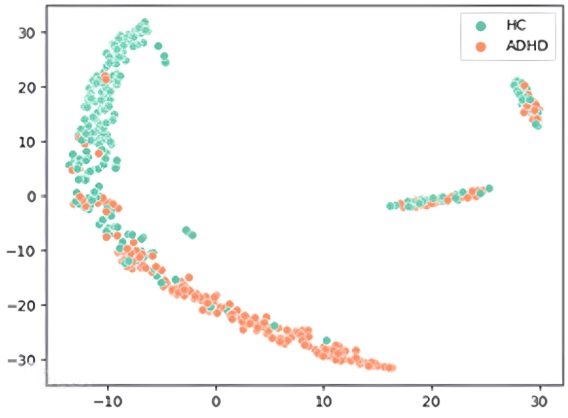}
		\caption{}
		\label{fig:six}
\end{subfigure}

	\caption{(a)Two-dimensional visualization obtained by t-SNE of the input features of ABIDE (b) and the output features of the first level (c) the second level.(d), (e), and (f) respectively represent similar operations on the ADHD-200.}
	\label{fig10}
\end{figure*}
\begin{figure*}[!h]
	\centering
	\begin{subfigure}{0.49\textwidth}
		\includegraphics[width=\textwidth]{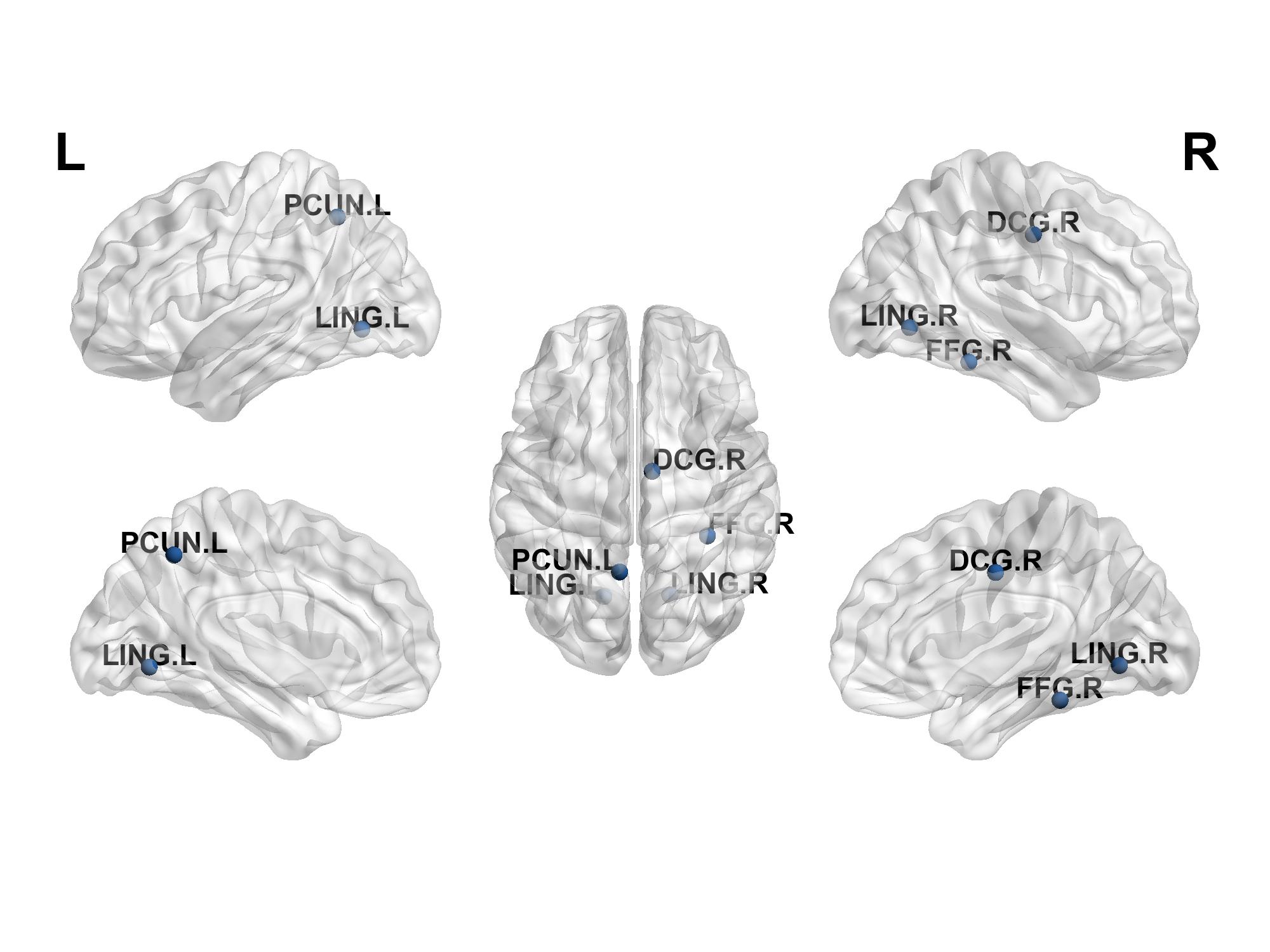}
		\caption{ABIDE}
		\label{fig:first2}
	\end{subfigure}
	\hfill
	\begin{subfigure}{0.49\textwidth}
		\includegraphics[width=\textwidth]{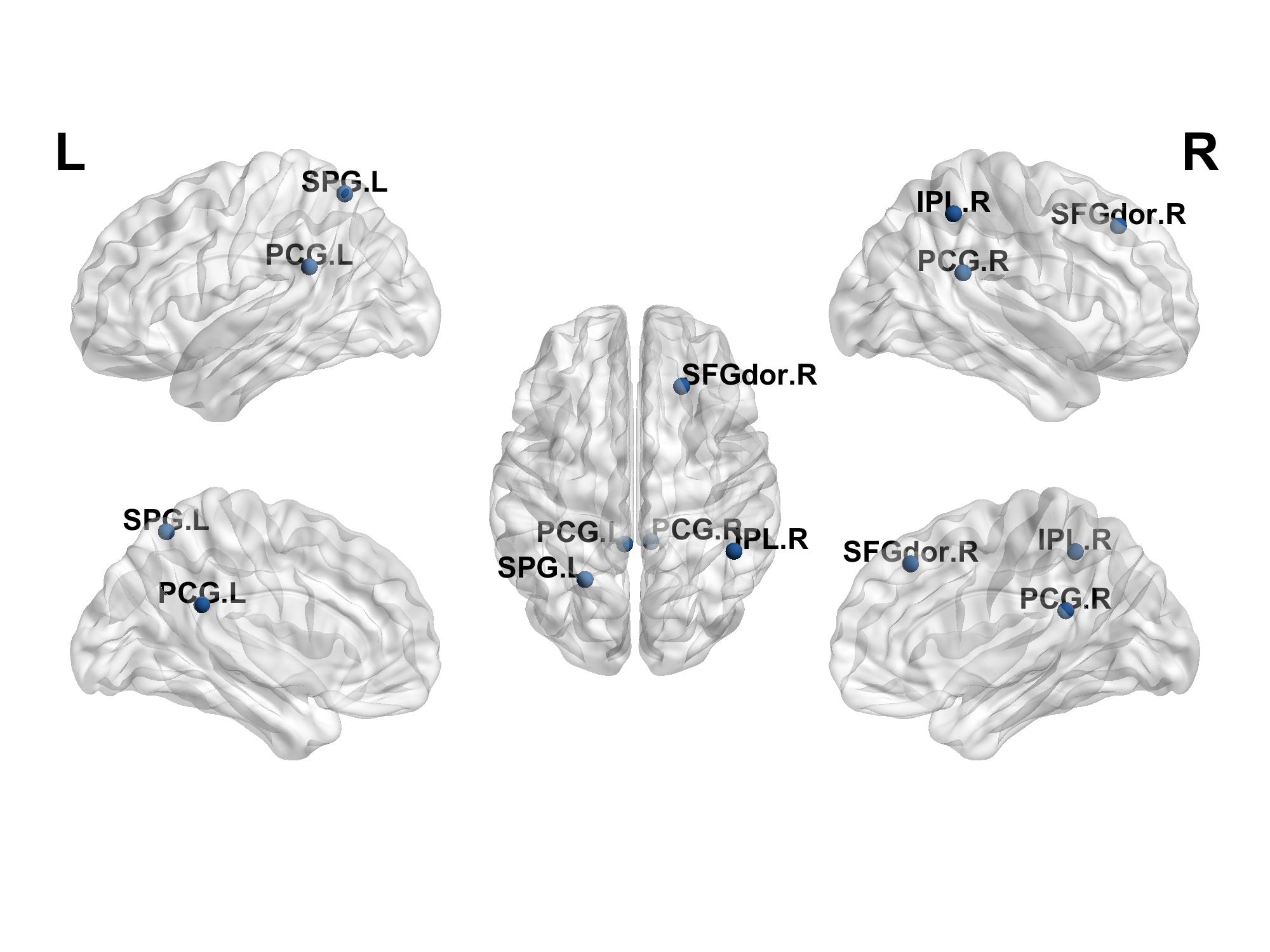}
		\caption{ADHD}
		\label{fig:second2}
	\end{subfigure}

	\caption{Identification of 5 important regions from the output of the SAGPool layer for the purpose of diagnosing brain disorders.}
	\label{fig11}
\end{figure*}
\subsection{Visualization of population graph effects}
As seen in the previous results, adding the population graph to the proposed model improves its performance. In this experiment, we attempted to visualize the impact of adding the population graph by representing the samples in a two-dimensional space using t-SNE. 
For each subject, we have a ROI-time series matrix $P\in R^{N\times T_{max}}$  that is converted into a feature vector by concatenating its rows before being fed as input to the model. Fig \ref{fig10}(a) and (d) depict these input samples, respectively, in dataset ABIDE and ADHD-200. As evident, the samples from two classes are completely mixed, and there is no clear boundary between them, indicating the difficulty in classification. Then, these samples are fed to the HDGL model, and the model is trained. Subsequently, the feature vectors obtained from the first-level (end of part (c)) in the last epoch shown in Fig \ref{fig10} (b) and (e) for datasets ABIDE and ADHD-200, respectively. In this scenario, the two classes, compared to the previous case, are separated from each other and positioned further apart. Subsequently, the feature vectors obtained from the second level (part (d)) of the model in the last epoch are illustrated in Fig \ref{fig10} (c) and (f), respectively, for the ABIDE and ADHD-200 datasets. As evident from the figures, the proposed model can clearly distinguish between the two classes and separate them further apart. This demonstrates the impact of the population graph in improving performance.

\subsection{Effective regions in diagnosing brain disorders}
In addition to classification, another goal of the proposed model design is interpretability of the results obtained from it. To this end, we designed an experiment to identify the regions that have played a more influential role in the diagnosis of ASD and ADHD from healthy samples. As mentioned earlier, SAGPool has the capability to identify and preserve more important nodes during training while discarding the rest. This module receives a sequence of $ T $ graphs over time, where each graph has $ N $ nodes representing $ N $ brain regions. By processing each of these graphs and removing some of their nodes, the module generates sequences of graphs in the output, each of which has  $\lceil kN \rceil $ nodes, where $ k $ represents the pooling ratio. Thus, after training HDGL using 5-fold cross-validation, the output graphs of the pooling layer are stored for the test datasets and examined. Assuming we have $ d $ test samples, each containing $ T $ graphs, each with $\lceil kN \rceil $ nodes, the output of the pooling layer generates $ d \times T \times \lceil kN \rceil $ nodes. Among these, we chose 5 frequently occurring nodes and introduced them as important regions in the diagnosis of brain disorders. This experiment was conducted for both ABIDE and ADHD-200 datasets, and 5 important regions are introduced in Fig\ref{fig11}, generated by using the BrainNet Viewer toolbox \citep{xia2013brainnet}. According to the this, the most influential regions in the diagnosis of autism are the Lingual cortex, consistent with findings from prior research \citep{gebauer2015there}. Other important regions include the Cingulate, Fusiform, and Precuneus cortices, which have shown associations with autism in studies such as \citep{cauda2011grey}. Additionally, the important regions identified for diagnosing ADHD are the Parietal, Cingulate, and Frontal regions, respectively. This aligns closely with the results presented in \citep{bush2011cingulate}.

\section{Limitations and Future Work}
\label{section6}
The proposed model determined the edge weights of the initial population graph using manually defined similarity metrics and a pre-determined function. This is in contrast to some recent studies where edge weights are learned through a trainable function, resulting in better performance. Incorporating these trainable functions into the proposed models may potentially lead to performance improvements.

Usually, in studies including the proposed model that employ sliding windows, it is necessary to standardize the lengths of time series samples before proceeding. For this purpose, a specific length is defined, and samples with lengths shorter or significantly longer than this value are often removed. On the other hand, fMRI are known for their high complexity, so the model requires a substantial amount of them to effectively learn these complexities. Removing samples can increase the risk of overfitting. To address this challenge, instead of removing samples,  upsampling techniques can be employed to standardize the lengths of time series.

We explored different combinations of phenotype features, including gender, age, and imaging site, to establish the edges of the population graph. To investigate the influence of additional features, one can leverage the data available in the datasets, such as handedness and IQ measurement, for constructing the population graph.

\section{Conclusion}
\label{section7}
In order to achieve a better graph embedding from brain networks, we developed a hierarchical dynamic graph representation learning model, called HDGL. The proposed HDGL model, after constructing brain network graphs and learning their spatial and temporal embeddings, models the population graph representing the connections between subjects and leverages their phenotype information. Furthermore, we introduced four training methods; three transductive methods and one inductive method. We demonstrated that one of transductive methods is scalable and not constrained by memory limitations. Extensive experiments were conducted to evaluate the effectiveness of our model on both ABIDE and ADHD-200 datasets, and the results indicated the improvement of our model compared to state-of-the-art models.






\bibliographystyle{unsrt}
\bibliography{refs}

\end{document}